\documentclass{article}

\usepackage[preprint]{neurips_2026}

\usepackage[utf8]{inputenc} %
\usepackage[T1]{fontenc}    %
\usepackage{hyperref}       %
\usepackage{url}            %
\usepackage{booktabs}       %
\usepackage{amsfonts}       %
\usepackage{nicefrac}       %
\usepackage{microtype}      %
\usepackage{xcolor}         %

\usepackage{enumitem}
\usepackage{amsmath} 
\usepackage{subcaption}
\usepackage{graphicx} 
\usepackage{amsthm}

\usepackage{multirow}
\usepackage{colortbl}
\usepackage{makecell}
\usepackage{array}
\usepackage{boldline}
\usepackage{placeins}
\usepackage{float}
\usepackage[most]{tcolorbox}

\newtheorem{analysis}{Theoretical Analysis}
\title{Future Forcing: Future-aware Training-free KV Cache Policy for Autoregressive Video Generation}

\author{%
\textbf{Jiayi Luo\textsuperscript{1,2} \quad
Qiyan Liu\textsuperscript{1} \quad
Tengyang Wang\textsuperscript{1} \quad
Junhao Liu\textsuperscript{1} \quad
Jiayu Chen\textsuperscript{3} \quad
Cong Wang\textsuperscript{4,2}} \\
\textbf{Hanxin Zhu\textsuperscript{5} \quad
Chen Gao\textsuperscript{1,6} \quad
Xiaobin Hu\textsuperscript{6} \quad
Qingyun Sun\textsuperscript{1,\textdagger} \quad
Zhibo Chen\textsuperscript{5,2,\textdagger}} \\[0.6em]
\textsuperscript{1} BUAA \quad
\textsuperscript{2} ZGCA \quad
\textsuperscript{3} PKU \quad
\textsuperscript{4} CASIA \quad
\textsuperscript{5} USTC \quad
\textsuperscript{6} NUS \\
\textsuperscript{\textdagger}Corresponding authors.
}
\newcommand{\modelname}{Future Forcing}


\begin{document}

\maketitle

\begin{abstract}
Autoregressive (AR) video generation has emerged as a promising paradigm for long-horizon video synthesis, where each frame is generated conditioned on previously generated tokens. 
To accelerate inference, the KV cache is used to avoid redundant recomputation across generation steps. 
Nevertheless, its growth with generation length introduces increasing memory and error accumulation, limiting the scalability of AR models to even longer sequences.
Existing KV cache compression methods mitigate this issue by selectively retaining only video tokens deemed important. 
However, most existing methods assess token importance using short-horizon signals derived from the current or historical generation context, making these methods prone to overlooking tokens that appear unimportant at early steps but later become critical for future frames.
In this work, we identify an important property of trained AR video models: \textit{\textbf{although RoPE-modulated queries evolve across autoregressive steps, the underlying canonical pre-RoPE query distribution remains remarkably stable throughout the video generation process}}.
This approximate stationarity implies that future query distributions are estimable from historical statistics, enabling principled future-aware cache decisions without any additional training.
Building on this insight, we propose \textbf{\underline{\modelname}}, a training-free future-aware KV cache policy for AR video generation.
Specifically, \modelname\ first constructs a future query proxy from historical statistics, then scores KV cache tokens by their importance under this proxy, and finally merges redundant token pairs within the affine subspace induced by the future query.
Extensive experiments show that \modelname\ improves long-horizon consistency under limited KV caches, achieving up to 1.49 improvement in subject consistency on VBench-Long for 60s generation over existing AR video KV cache policies.
\end{abstract}

\section{Introduction}

Bidirectional Diffusion Transformer (DiT)-based video generation models~\cite{wan2025wan,kong2024hunyuanvideo,wu2025hunyuanvideo,chen2025hunyuanvideo,hong2022cogvideo} employ bidirectional attention to jointly denoise all frames, enabling coherent full-sequence modeling and high-quality synthesis. Yet this paradigm is inherently offline, ill-suited for streaming or continual video extension~\cite{huang2025self,yin2025causvid}. By contrast, autoregressive (AR) video models~\cite{zhu2026causalforcing,liu2025rolling} causally condition each frame on prior tokens, enabling efficient incremental generation.

To make incremental decoding efficient, AR video models reuse historical key and value representations through KV caching, avoiding repeated computation over the full history at each step. Since the cache grows with generation length, practical memory constraints require operating under a bounded KV cache budget. Under this constraint, cache selection becomes critical: evicting tokens needed by future frames distorts attention over historical context, and these errors propagate through subsequent autoregressive steps, accumulating into visual degradation, temporal inconsistency, and motion drift.

However, most existing KV cache compression methods for AR video generation estimate token importance from current or recent context-derived signals~\cite{yi2025deep,zhu2026causalforcing,liu2025rolling,lu2025rewardforcing,huang2025self}. These signals reflect present relevance but cannot anticipate future attention demands, as future query representations are unavailable at KV cache eviction time. \textit{\textbf{This induces a short horizon bias under a limited cache budget}}: \textit{tokens with low current importance may be discarded before becoming critical for scene structure, object identity, or motion continuity, causing attention errors that accumulate into severe long horizon degradation.}

In this work, we identify an intriguing property of trained AR video models by examining query representations before and after RoPE modulation. In transformer attention, rotary positional embedding (RoPE)~\cite{su2021roformer} is applied to queries before attention weights are computed, producing position encoded queries that determine how each frame attends to cached history. Our key observation is that \textbf{these RoPE-modulated attention queries vary substantially across autoregressive steps, whereas their underlying pre-RoPE query distributions remain stable under a fixed prompt}. This reveals a separation between stable semantic query structure and step dependent positional modulation, providing a basis for anticipating future attention behavior without additional training.

Building on this insight, we propose \modelname, a future-aware training-free KV cache policy for AR video generation that preserves generation quality and consistency under a fixed cache budget to alleviate short-horizon limitations. 
Specifically, \modelname\ first constructs query proxies based on the stable pre-RoPE query distribution and uses them to score cached tokens by their predicted future importance, rather than relying only on their current or historical importance.
Then, our \modelname\ further merges evicted tokens into retained entries with similar future-query-induced attention profiles to reduce compression-induced loss. In summary, our core contributions are as follows:

\begin{figure}[t]
    \centering
    \includegraphics[width=\linewidth]{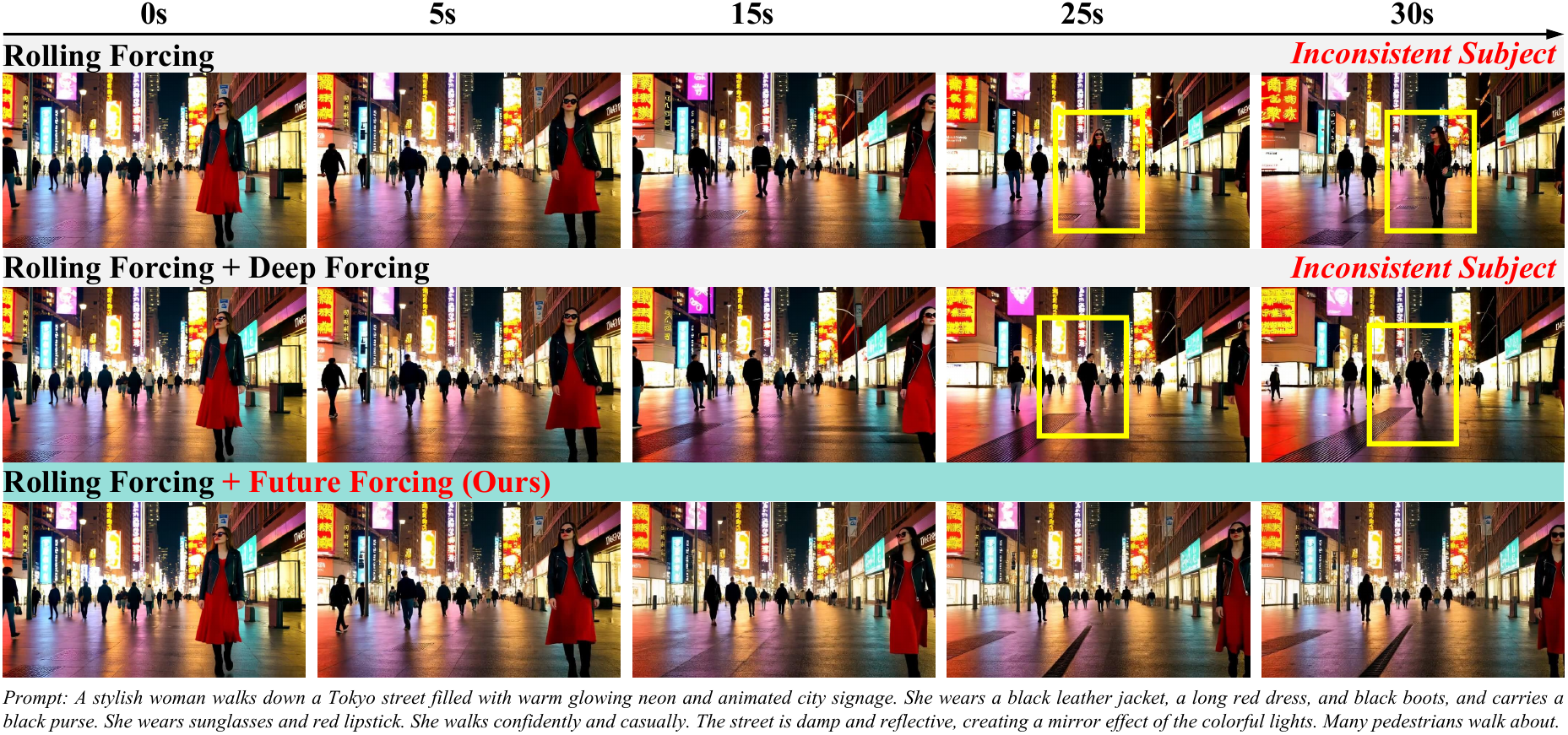}
    \caption{Qualitative comparison of 30s long video generation across different KV cache strategies. Competing methods exhibit temporal and subject inconsistency as generation progresses, while our \modelname\ maintains coherent subject consistency throughout the full AR generation process.}
    \label{fig:intro}
\end{figure}

\begin{itemize}[leftmargin=*]
    \item We investigate query representation dynamics in well-trained AR video generation models and uncover an intriguing phenomenon: RoPE-modulated queries vary substantially across autoregressive steps, whereas the underlying pre-RoPE query distribution remains stable under a fixed prompt.
    
    \item We investigate the short-horizon limitation of existing KV cache compression methods, which rely on local context-derived signals and may discard tokens that are critical for future decoding. To address this, we propose \modelname, a future-aware training-free KV cache policy that uses future query proxies to guide token eviction and merging under the fixed KV cache budget.
    
    \item Extensive experiments on representative autoregressive video generation models show that \modelname\ outperforms existing autoregressive video KV cache policies, achieving up to 1.49 subject-consistency improvement while maintaining generation quality under limited KV cache budgets.
\end{itemize}

\section{Related Work}
\subsection{Bidirectional Video Generation Models.}
Recent progress in video generation has been largely driven by bidirectional diffusion-transformer models that jointly denoise entire video clips in spatiotemporal latent space. Sora~\cite{brooks2024video} showed that scaling DiT pre-training on spacetime patches yields high-fidelity and temporally coherent videos, inspiring many follow-up systems. Open-source models including CogVideoX~\cite{yang2024cogvideox,hong2022cogvideo}, HunyuanVideo~\cite{kong2024hunyuanvideo,wu2025hunyuanvideo,chen2025hunyuanvideo}, Wan~\cite{wan2025wan}, Open-Sora~\cite{zheng2024opensora}, Latte~\cite{ma2024latte}, LTX-Video~\cite{hacohen2024ltxvideo}, Mochi-1~\cite{genmo2024mochi}, Step-Video-T2V~\cite{ma2025stepvideo}, and Pyramid Flow~\cite{jin2024pyramidal} have substantially improved visual fidelity, motion quality, and text-video alignment. Proprietary systems such as Movie Gen~\cite{polyak2024moviegen}, Veo~\cite{deepmind2024veo}, and Kling~\cite{kuaishou2024kling} follow similar design principles at even larger scales. \textit{However}, jointly denoising all spatiotemporal tokens before decoding any frame precludes streaming generation during inference, making these methods ill-suited for real-time, low-latency interactive applications.

\subsection{Autoregressive Video Generation Models.}

Autoregressive video generation models synthesize videos sequentially from previously generated context, making them naturally suited to streaming and long-horizon extension. 
Early works like VideoGPT~\cite{yan2021videogpt}, Phenaki~\cite{villegas2023phenaki}, VideoPoet~\cite{kondratyuk2023videoPoet}, and Genie~\cite{bruce2024genie} established token-based and variable-length video generation. More recent methods, including Progressive AR Video Diffusion Models~\cite{xie2024progressive}, AR-Diffusion~\cite{sun2025ardiffusion}, NOVA~\cite{deng2024nova}, VideoMAR~\cite{yu2025videomar}, MAGI~\cite{zhou2025magi}, and VideoAR~\cite{ji2026videoar}, further strengthen this paradigm through spatiotemporal factorization and causal diffusion modeling. Recent advances such as CausVid~\cite{yin2025causvid}, Self Forcing~\cite{huang2025self}, Causal Forcing~\cite{zhu2026causalforcing}, Rolling Forcing~\cite{liu2025rolling}, Reward Forcing~\cite{lu2025rewardforcing}, Sparse Forcing~\cite{xu2026sparse}, LongLive~\cite{yang2025longlive}, and Context Forcing~\cite{chen2026contextforcing} push AR video generation toward stable, interactive, and long-horizon synthesis, while MemoryPack~\cite{wu2025memorypack}, VideoSSM~\cite{yu2025videossm}, HiAR~\cite{zou2026hiar}, FLEX~\cite{li2026flex}, and Helios~\cite{yuan2026helios} further explore efficient modeling.
\textit{However}, during long video generation, AR models typically rely on KV cache to avoid recomputing prior context, which increases GPU memory overhead and further aggravates the error accumulation.

\subsection{KV Cache in AR Video Generation Models}

KV-cache compression has been extensively studied in autoregressive LLM inference, with representative methods including Scissorhands~\cite{scissorhands}, H$_2$O~\cite{h2o}, and PyramidKV~\cite{pyramidkv}. 
In autoregressive video generation, Self Forcing~\cite{huang2025self} introduces a rolling KV cache, while LongLive~\cite{yang2025longlive} combines KV-recache, short-window attention, and frame sinks for interactive long-video synthesis. More direct cache-centric approaches include PackCache~\cite{li2026packcache}, Dummy Forcing~\cite{dummyforcing}, FlowCache~\cite{ma2026flow}, TempCache in Fast AR Video Diffusion~\cite{fastarvideo}, Quant VideoGen~\cite{quantvideogen}, and PaFu-KV~\cite{chen2026past}, which respectively explore token compaction, and low-bit KV quantization, and salience-based eviction. Recent works such as Relax Forcing~\cite{relaxforcing}, DeepForcing~\cite{yi2025deep}, Anchor Forcing~\cite{anchorforcing}, and PackForcing~\cite{packforcing} further organize generation history into structured memory regions, while a recent 33-method empirical study highlights the strong impact of cache policy on VRAM, runtime, and long-horizon fidelity in Self-Forcing-style generation~\cite{kvquantstudy}. \textit{However, existing methods typically rely on short-horizon signals from the current or historical generation context to make cache decisions, which can overlook tokens that seem unimportant early but later become critical for future frames.}

\section{Motivation and Analysis}
\subsection{Preliminaries}

\textbf{Autoregressive Video Generation and KV Caching.}
We consider autoregressive (AR) video generation in the latent space~\cite{huang2025self}. Given a prompt $c$, the model generates a video token sequence $\mathbf{x}_{1:T}=[\mathbf{x}_1,\ldots,\mathbf{x}_T]$ causally, where $\mathbf{x}_t \in \mathbb{R}^{N\times d}$ denotes the $N$ latent tokens of frame $t$. The generation process is factorized as \( p(\mathbf{x}_{1:T}\mid c)=\prod_{t=1}^{T} p(\mathbf{x}_t \mid \mathbf{x}_{<t}, c) \), so each frame is predicted conditioned only on previously generated context. For a token $i$ with hidden representation $\mathbf{h}_i\in\mathbb{R}^d$, its query, key, and value are defined as \( \mathbf{q}_i=\mathbf{W}_Q\mathbf{h}_i \), \( \mathbf{k}_i=\mathbf{W}_K\mathbf{h}_i \), and \( \mathbf{v}_i=\mathbf{W}_V\mathbf{h}_i \). For a newly decoded token, causal self-attention is computed over previously generated tokens, with attention weights \( \alpha_{ij}=\frac{\exp(\tilde{\mathbf{q}}_i^\top \tilde{\mathbf{k}}_j/\sqrt{d})}{\sum_{\ell\le i}\exp(\tilde{\mathbf{q}}_i^\top \tilde{\mathbf{k}}_\ell/\sqrt{d})} \) and output \( \mathrm{Attn}(\mathbf{q}_i,\mathbf{K}_{\le i},\mathbf{V}_{\le i})=\sum_{j\le i}\alpha_{ij}\mathbf{v}_j \), where $\tilde{\mathbf{q}}_i$ and $\tilde{\mathbf{k}}_j$ denote the position-modulated query and key. During autoregressive decoding, past keys and values are typically cached and reused to avoid recomputing the entire history during generation.

\textbf{RoPE in AR Video Generation.}
To encode spatiotemporal positions, Rotary Position Embedding (RoPE) is commonly adopted in transformer attention layers for video generation~\cite{su2021roformer,wei2025videorope,liu2025vrope}. For a token at 3D position $\mathbf{p}_i=(t_i,y_i,x_i)$, RoPE applies a position-dependent transformation to the query and key, in its standard matrix notation, as \( \tilde{\mathbf{q}}_i = R(\mathbf{p}_i)\mathbf{q}_i \) and \( \tilde{\mathbf{k}}_i = R(\mathbf{p}_i)\mathbf{k}_i \), where $R(\mathbf{p}_i)$ denotes the corresponding RoPE transformation determined by the token's spatiotemporal coordinates. Under this notation, the attention score can be explicitly written as \( \tilde{\mathbf{q}}_i^\top \tilde{\mathbf{k}}_j=\mathbf{q}_i^\top R(\mathbf{p}_i)^\top R(\mathbf{p}_j)\mathbf{k}_j \), directly showing that RoPE injects positional information through the interaction between the position-dependent transformations applied to queries and keys. In this work, we refer to $\mathbf{q}_i$ as the \textit{canonical pre-RoPE query} and to $\tilde{\mathbf{q}}_i$ as the \textit{RoPE-modulated query}. The former captures the underlying content-dependent query representation before positional modulation, while the latter is the query actually used in attention computation during autoregressive decoding.

\begin{figure}[t]
    \centering
    \begin{subfigure}[t]{0.495\linewidth}
        \centering
        \includegraphics[width=\linewidth]{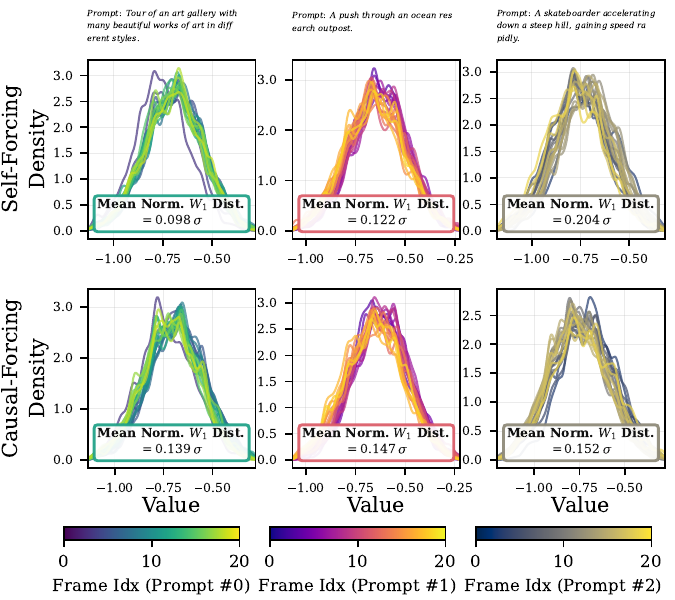}
        \caption{Pre-RoPE queries distribution (\textbf{Stable}).}
        \label{fig:pre_rope_query}
    \end{subfigure}
    \hfill
    \begin{subfigure}[t]{0.495\linewidth}
        \centering
        \includegraphics[width=\linewidth]{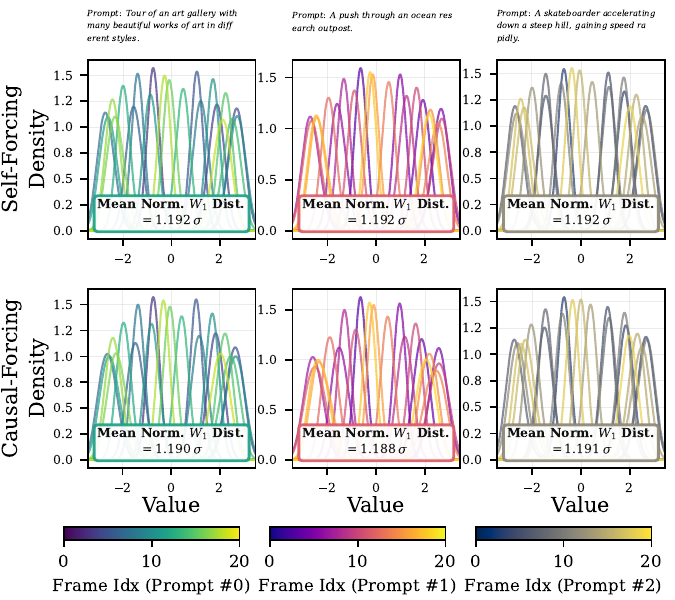}
        \caption{Post-RoPE queries distribution (\textbf{Varying}).}
        \label{fig:rope_query}
    \end{subfigure}
    \caption{Query distribution across latent frames for different autoregressive video generation models. (Quantitative results based on mean normalized Wasserstein distance are detailed in Appendix~\ref{sec:appendix_stability_quantitative}). While the distributions of \textit{RoPE-modulated queries vary noticeably} across autoregressive generation stages, the distributions of the corresponding \textit{pre-RoPE queries remain considerably more stable}.}
    \vspace{-1em}
    \label{fig:query_distribution}
\end{figure}

\subsection{\textit{Motivation}: Pre-RoPE Query Distribution Stability Across AR Video Generation}
\label{sec:emprical_analysis}
In this section, we investigate how query representations evolve throughout autoregressive video generation. Specifically, we consider two representative autoregressive video generation models, namely Self-Forcing and Causal-Forcing, and visualize the distributions of both canonical pre-RoPE queries and RoPE-modulated queries across latent frames at different autoregressive generation stages using three randomly selected prompts. 
Additional visualizations, including highly dynamic scenarios, are provided in Appendices~\ref{sec:appendix_exp} and~\ref{sec:highly_dynamic_appendix}.
As shown in Figure~\ref{fig:query_distribution}, \textbf{under a fixed prompt, the distributions of RoPE-modulated queries vary noticeably across different autoregressive generation stages, whereas the corresponding distributions of pre-RoPE queries remain considerably more stable over time}. 
This consistent empirical phenomenon suggests that the apparent temporal drift in query representations is primarily introduced by RoPE-based positional modulation, while the underlying canonical query distribution remains relatively stable over time under a fixed prompt, thereby enabling the future attention behavior to be more reliably derived from the historical pre-RoPE query statistics.

\section{Method}
\label{sec:method}

\begin{figure}[t]
    \centering
    \includegraphics[width=\linewidth]{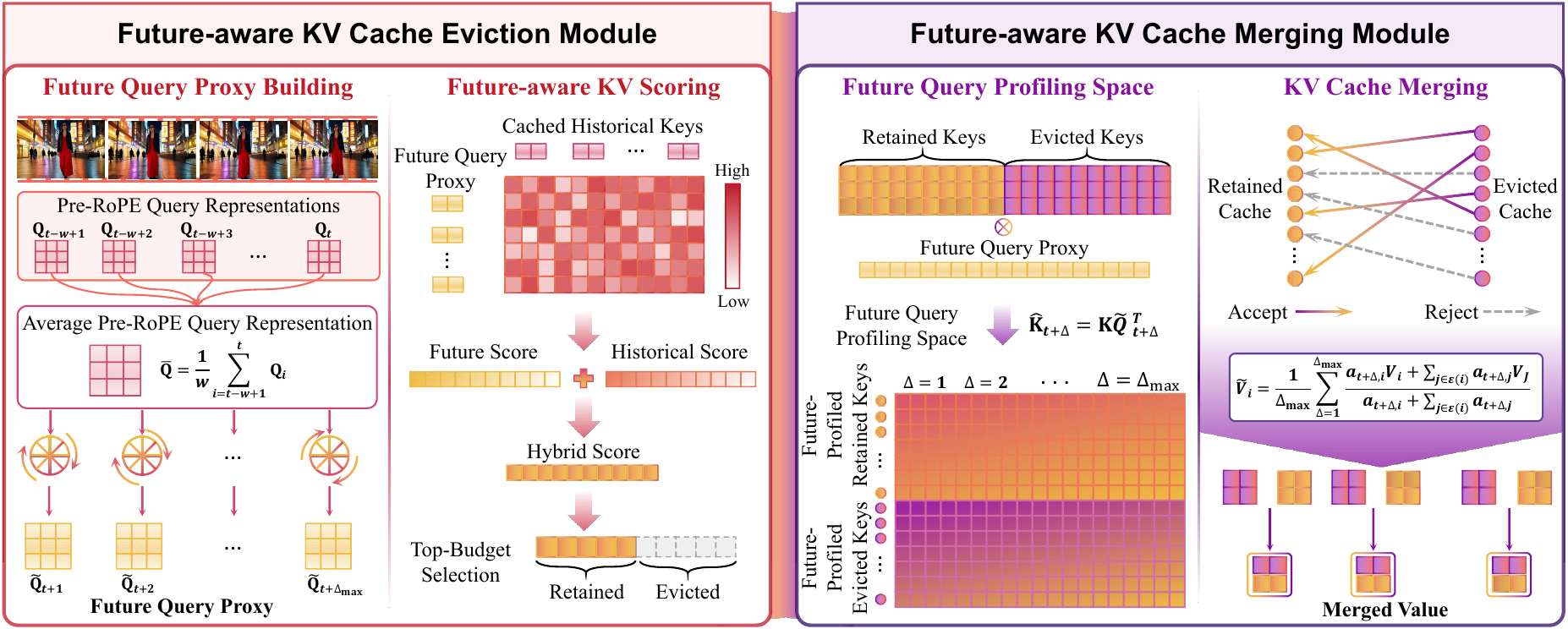}
    \caption{Overview of \modelname, which constructs future query proxies from stable pre-RoPE queries, combines future- and history-aware scores to retain KV pairs, and merges evicted entries into retained ones in future-query profiling space, reducing information loss under a limited cache budget.}
    \label{fig:framework}
\end{figure}

In this section, we introduce our \textbf{\underline{\modelname}}, a future-aware training-free KV cache policy for autoregressive video generation that reduces KV-cache memory overhead while mitigating the error accumulation during long-horizon decoding. As illustrated in Figure~\ref{fig:framework}, \modelname\ comprises two complementary modules: (1) a \textit{Future-aware KV Cache Eviction Module} that estimates future query proxies from historical pre-RoPE statistics to retain only the most relevant KV-cache entries; and (2) a \textit{Future-aware KV Cache Merging Module} that consolidates evicted entries into the retained cache in the future query profiling space, minimizing the information loss under a fixed KV budget.

\subsection{Future-aware KV Cache Eviction Module}
As shown by the empirical analysis in Figure~\ref{fig:query_distribution} of Section~\ref{sec:emprical_analysis}, although RoPE-modulated queries vary noticeably across autoregressive steps, the distributions of pre-RoPE queries remain relatively stable under fixed prompts. Intuitively, under a fixed prompt, the model maintains a consistent underlying semantic focus throughout the entire generation process. What changes across autoregressive steps is not this global semantic focus itself, but rather its positional instantiation: RoPE maps the canonical query direction into different spatiotemporal configurations at each generation step, thereby yielding step-dependent local attention patterns while leaving the pre-RoPE query distribution largely intact overall. Building on this empirical observation, we now provide a concise theoretical explanation:

\begin{analysis}[Query Drift Bound]\label{thorem:upper_bound}
Given a fixed prompt $c$, for a existing trained forcing-based AR video generation transformer $f_{\boldsymbol{\theta}}$, let $\mathbf{X}^t \in \mathbb{R}^{N \times d}$ denote the latent representation at step $t$, and let $\mathbf{Q}^t = \mathbf{X}^t \mathbf{W}_Q^\top$ and $\tilde{\mathbf{Q}}^t$ denote the pre-RoPE and RoPE-modulated query matrices, respectively. We measure query drift via relative change $\rho_{\mathrm{pre}}^{t \to s} = \|\mathbf{Q}^s - \mathbf{Q}^t\|_F / \|\mathbf{Q}^t\|_F$ and $\rho_{\mathrm{post}}^{t \to s} = \|\tilde{\mathbf{Q}}^s - \tilde{\mathbf{Q}}^t\|_F / \|\tilde{\mathbf{Q}}^t\|_F$. Let $\epsilon = \|\mathbf{X}^s - \mathbf{X}^t\|_F$ denote the latent difference between steps, $\Delta p = p_t - p_s$ the relative positional offset, and $\theta$ the RoPE base frequency. The following two upper bounds hold:
\begin{equation}
\rho_{\mathrm{pre}}^{t\to s} \leq \frac{\|\mathbf{W}_Q\|_{\mathrm{op}} \cdot \epsilon}{\|\mathbf{Q}^t\|_F}, \qquad
\rho_{\mathrm{post}}^{t\to s} \leq 2\left|\sin\frac{\Delta p \cdot \theta}{2}\right| + \frac{\|\mathbf{W}_Q\|_{\mathrm{op}} \cdot \epsilon}{\|\mathbf{Q}^t\|_F}.
\end{equation}
\end{analysis}

Theoretical Analysis~\ref{thorem:upper_bound} characterizes upper bounds on the relative drift of pre-RoPE and RoPE-modulated queries, with proofs in Appendix~\ref{sec:appendix_proof}. In practice, since $N\gg d'$ makes $\|\mathbf{Q}^t\|_F$ accumulate over many tokens and typically yields $\|\mathbf{W}_Q\|_{\mathrm{op}} \ll \|\mathbf{Q}^t\|_F$, while frames remain highly redundant under a fixed prompt~\cite{quantvideogen,fastarvideo,ma2026flow}, $\epsilon$ stays small across many step offsets, making the bound on $\rho_{\mathrm{pre}}^{t\to s}$ consistently \textit{tight}. By contrast, for standard RoPE, the lowest frequency $\theta = 10000^{-2\cdot 0/128} = 1$~\cite{huang2025self} ensures that even for adjacent frames with $\Delta p = 1$, the positional term $2\left|\sin\frac{\Delta p\cdot\theta}{2}\right| \approx 0.958$ remains large regardless of $\epsilon$, making the bound on $\rho_{\mathrm{post}}^{t\to s}$ substantially \textit{looser}, consistent with the instability patterns observed in Figure~\ref{fig:query_distribution} Section~\ref{sec:emprical_analysis}.

Building on the theoretical and empirical analyses of pre-RoPE query stability, we estimate future attention behavior without any additional forward passes by constructing RoPE-modulated query proxies from the stable canonical distribution. Formally, let $\{\mathbf{Q}_{t-w+1},\dots,\mathbf{Q}_t\}$ denote the pre-RoPE query matrices observed over the most recent $w$ generated frames. Since the pre-RoPE query distribution is stable across time, their mean $\bar{\mathbf{Q}} = \frac{1}{w}\sum_{s=t-w+1}^{t} \mathbf{Q}_s$ serves as a time-invariant canonical representative. For any future frame $t+\Delta$, where $\Delta \in \{1,\dots,\Delta_{\max}\}$, we construct a RoPE-modulated query proxy $\tilde{\mathbf{Q}}_{t+\Delta} = \mathcal{R}_{t+\Delta}(\bar{\mathbf{Q}})$, where $\mathcal{R}_{t+\Delta}(\cdot)$ denotes the RoPE positional encoding operator applied at position $t+\Delta$. This proxy approximates the positionally-encoded query that frame $t+\Delta$ would issue without requiring it to be generated. 
We then estimate the attention logit of frame $t+\Delta$ against each cached key $\hat{\mathbf{K}}_j$ as $\hat{s}_{t+\Delta,j} = \langle \tilde{\mathbf{Q}}_{t+\Delta},\, \hat{\mathbf{K}}_j \rangle / \sqrt{d_h}$, and obtain its softmax-normalized weight $\hat{a}_{t+\Delta,j}$ as a per-token future importance estimate. The final cache entry score combines predicted future relevance with average historical attention across frames as follows:
\begin{equation}\label{eq:score}
    \mathrm{score}_j
    \;=\;
    \lambda \underbrace{\frac{1}{\Delta_{\max}}\sum_{\Delta=1}^{\Delta_{\max}} \hat{a}_{t+\Delta,j}}_{\text{future}}
    \;+\;
    (1-\lambda)\underbrace{\frac{1}{t}\sum_{s=1}^{t} a_{s,j}}_{\text{history}},
\end{equation}
where $\lambda$ is a hyperparameter balancing future relevance and historical attention. Cache entries are ranked by this score, and those falling below the budget threshold are evicted from the KV cache.

\textbf{Custom Attention Kernel Design.}
Since existing attention kernels such as FlashAttention~\citep{dao2022flashattention,dao2023flashattention2} do not expose per-token attention scores, we implement a lightweight two-pass Triton kernel to compute Eq.~\eqref{eq:score} without materializing the full $\mathbf{Q}\mathbf{K}^\top$ attention score matrix, reducing future-aware scoring computational overhead. Implementation details are provided in Appendix~\ref{sec:appendix_kernel}.

\subsection{Future-aware KV Cache Merging Module}
Under a fixed KV cache budget, naively discarding evicted entries introduces irreversible information loss. A straightforward remedy is to merge entries with high key similarity~\citep{zhang2024cam}, so that evicted information is absorbed into retained ones. However, as the attention output $\mathbf{O} = \sum_j a_j \mathbf{V}_j$ depends on the attention weights $a_j = \mathrm{softmax}(\mathbf{Q}\mathbf{K}_j^\top/\sqrt{d_h})$, \textit{the contribution of each KV entry is inherently query-dependent}: two close keys may receive drastically different attention weights under different queries, making key-similarity alone an unreliable merging criterion. 
Instead, we propose to perform merging in the future-query profiling space: rather than measuring proximity by Euclidean distance, we group entries by the similarity of their predicted attention weight profiles under future queries, directly minimizing attention output distortion rather than treating KV independently.

Specifically, let $\mathcal{S}$ and $\mathcal{E}$ denote the retained and evicted index sets. For each evicted entry $j\in\mathcal{E}$, we determine its merge target by measuring similarity in the future query profiling space. Projecting all keys into this space via $\hat{\mathbf{K}}_{t+\Delta} = \mathbf{K}\tilde{\mathbf{Q}}_{t+\Delta}^{\top}\in\mathbb{R}^{N\times\Delta_{\max}}$, high cosine similarity between $\hat{\mathbf{K}}_i$ and $\hat{\mathbf{K}}_j$ indicates that entries $i\in\mathcal{S}$ and $j\in\mathcal{E}$ attract similar future attention and are therefore safe to merge. We route each evicted entry to its nearest retained neighbor $i^\star(j) = \arg\max_{i\in\mathcal{S}}\cos(\hat{\mathbf{K}}_i, \hat{\mathbf{K}}_j)$, and admit the merge only when $\cos(\hat{\mathbf{K}}_{i^\star(j)}, \hat{\mathbf{K}}_j)\ge\tau$ for fidelity threshold $\tau\in[0,1]$, discarding $j$ otherwise. The merged key is set to $\widetilde{\mathbf{K}}_i = \mathbf{K}_i$ to preserve RoPE consistency, while the merged value is obtained by weighting each entry by its future attention mass across the look-ahead horizon:
\begin{equation}
\widetilde{\mathbf{V}}_i
\;=\;
\frac{1}{\Delta_{\max}}\sum_{\Delta=1}^{\Delta_{\max}}
\frac{\hat{a}_{t+\Delta,i}\,\mathbf{V}_i + \sum_{j\in\mathcal{E}(i)} \hat{a}_{t+\Delta,j}\,\mathbf{V}_j}{\tilde{a}_{t+\Delta,i}},
\end{equation}
where $\tilde{a}_{t+\Delta,i} = \hat{a}_{t+\Delta,i} + \sum_{j\in\mathcal{E}(i)}\hat{a}_{t+\Delta,j}$ is the re-normalized attention mass at slot $i$, and $\mathcal{E}(i)=\{j\in\mathcal{E}: i^\star(j)=i,\;\cos(\hat{\mathbf{K}}_i,\hat{\mathbf{K}}_j)\ge\tau\}$ is the set of admitted evicted entries routed to $i$.

\textbf{Dynamic RoPE in Long Video Generation.}
Following~\cite{yi2025deep,zhu2026causalforcing}, as absolute frame indices grow beyond the model's trained frequency range during long video generation, we apply a RoPE correction after each eviction step that remaps surviving cached frame positions to be contiguous immediately before the current generation position. This correction is applied in-place to the time-axis of cached key tensors, leaving spatial RoPE channels untouched, ensuring relative query-key temporal distances remain within the model's trained distribution throughout generation.

\section{Experiment}

\definecolor{mygreen}{RGB}{151,223,217}

\newcommand{\gc}[1]{\cellcolor{mygreen!50!white}#1}
\newcommand{\gb}[1]{\cellcolor{mygreen!50!white}\textbf{#1}}
\newcommand{\gmethod}[1]{\cellcolor{mygreen!50!white}\textbf{#1}}

\begin{table*}[t]
\centering
\caption{Quantitative comparison of \modelname\ and baselines across \textbf{30-second and 60-second long video generation} on VBenchLong~\cite{huang2024vbench}, using MovieGen prompts~\cite{polyak2024moviegen}. $\uparrow$ denotes higher is better for all metrics. The best results within each group are bold.}
\label{tab:long-video}
\setlength{\tabcolsep}{1pt}
\renewcommand{\arraystretch}{1.15}
\resizebox{\textwidth}{!}{%
\begin{tabular}{c|l|ccccccc}
\toprule
\textbf{Config} & \multicolumn{1}{c|}{\textbf{Backbone + KV Cache Policy}}
  & \makecell{\textbf{Subject}\\\textbf{Consist.}$\uparrow$}
  & \makecell{\textbf{Background}\\\textbf{Consist.}$\uparrow$}
  & \makecell{\textbf{Motion}\\\textbf{Smooth.}$\uparrow$}
  & \makecell{\textbf{Dynamic}\\\textbf{Degree}$\uparrow$}
  & \makecell{\textbf{Aesthetic}\\\textbf{Quality}$\uparrow$}
  & \makecell{\textbf{Imaging}\\\textbf{Quality}$\uparrow$}
  & \makecell{\textbf{Overall}\\\textbf{Consist.}$\uparrow$} \\
\midrule
\multirow{12}{*}{30s}
  & Self Forcing+Native                   & 96.01 & 95.20 & 98.16 & 45.27 & 55.43 & 68.92 & 23.09 \\
  & Self Forcing+Deep Forcing             & 96.39 & 95.75 & 98.36 & 59.34 & 57.67 & 69.05 & \textbf{23.91} \\
  & \gmethod{Self Forcing+\modelname}     & \gb{97.22} & \gb{96.18} & \gb{98.48} & \gb{63.08} & \gb{57.93} & \gb{70.22} & \gc{23.60} \\
\cmidrule(l){2-9}
  & Reward Forcing+Native                 & 97.02 & \textbf{96.06} & 98.34 & \textbf{70.28} & 57.04 & 70.29 & \textbf{23.94} \\
  & Reward Forcing+Deep Forcing           & 96.81 & 96.03 & 98.35 & 57.62 & 56.80 & 68.11 & 23.75 \\
  & \gmethod{Reward Forcing+\modelname}   & \gb{97.11} & \gc{95.96} & \gb{98.57} & \gc{69.77} & \gb{57.68} & \gb{70.35} & \gc{23.61} \\
\cmidrule(l){2-9}
  & Rolling Forcing+Native                & 97.91 & 96.56 & 98.59 & 34.20 & 57.15 & 70.89 & 23.25 \\
  & Rolling Forcing+Deep Forcing          & 97.87 & 96.54 & 98.78 & \textbf{35.97} & 57.68 & 70.29 & 23.84 \\
  & \gmethod{Rolling Forcing+\modelname}  & \gb{98.15} & \gb{96.72} & \gb{98.81} & \gc{32.42} & \gb{58.04} & \gb{71.48} & \gb{24.01} \\
\cmidrule(l){2-9}
  & Causal Forcing+Native                 & 94.74 & 94.70 & 96.80 & 97.14 & 58.71 & \textbf{69.07} & 23.85 \\
  & Causal Forcing+Deep Forcing           & 94.64 & 94.65 & 96.75 & \textbf{97.23} & 58.49 & 68.85 & 23.87 \\
  & \gmethod{Causal Forcing+\modelname}   & \gb{96.11} & \gb{95.45} & \gb{97.47} & \gc{88.16} & \gb{59.31} & \gc{\textbf{69.07}} & \gb{24.51} \\
\midrule
\multirow{12}{*}{60s}
  & Self Forcing+Native                   & 96.45 & 95.12 & 98.46 & 36.15 & 54.23 & 67.28 & 20.30 \\
  & Self Forcing+Deep Forcing             & 96.42 & \textbf{96.09} & 98.45 & 51.01 & \textbf{57.47} & 68.86 & 23.80 \\
  & \gmethod{Self Forcing+\modelname}     & \gb{97.01} & \gc{96.06} & \gb{98.54} & \gb{62.63} & \gc{57.21} & \gb{69.94} & \gb{23.87} \\
\cmidrule(l){2-9}
  & Reward Forcing+Native                 & 96.50 & 95.47 & 98.30 & \textbf{72.35} & 57.74 & 69.84 & \textbf{23.94} \\
  & Reward Forcing+Deep Forcing           & \textbf{97.08} & 95.14 & 98.22 & 60.45 & 56.68 & 66.62 & 23.37 \\
  & \gmethod{Reward Forcing+\modelname}   & \gc{96.55} & \gb{95.78} & \gb{98.59} & \gc{69.12} & \gb{58.71} & \gb{69.99} & \gc{23.73} \\
\cmidrule(l){2-9}
  & Rolling Forcing+Native                & 97.73 & 96.48 & 98.76 & 34.20 & 56.65 & 71.09 & 23.25 \\
  & Rolling Forcing+Deep Forcing          & 97.69 & 96.41 & 98.75 & \textbf{34.97} & 56.87 & 71.03 & 23.19 \\
  & \gmethod{Rolling Forcing+\modelname}  & \gb{98.05} & \gb{96.60} & \gb{98.78} & \gc{31.43} & \gb{57.71} & \gb{71.38} & \gb{23.50} \\
\cmidrule(l){2-9}
  & Causal Forcing+Native                 & 94.62 & 94.61 & 96.71 & \textbf{97.40} & 58.30 & 68.94 & 23.50 \\
  & Causal Forcing+Deep Forcing           & 94.50 & 94.57 & 96.67 & 97.30 & 58.08 & 68.62 & 23.57 \\
  & \gmethod{Causal Forcing+\modelname}   & \gb{96.11} & \gb{95.44} & \gb{97.48} & \gc{86.79} & \gb{59.09} & \gb{69.31} & \gb{24.20} \\
\bottomrule
\end{tabular}%
}
\end{table*}

\subsection{Experimental Settings}
\textbf{Implementation Details.}
We evaluate our \modelname\ across a diverse set of representative autoregressive video diffusion generation models, including Self Forcing~\cite{huang2025self}, Causal Forcing~\cite{zhu2026causalforcing}, Rolling Forcing~\cite{liu2025rolling}, Reward Forcing~\cite{lu2025rewardforcing}, LongLive~\cite{yang2025longlive}, and Deep Forcing~\cite{yi2025deep}. For the main hyperparameters, we set the future-history balance $\lambda=0.5$, look-ahead horizon $\Delta_{\max}=6$, and merge threshold $\tau=0.95$. For the used KV cache budget, we retain 18 latent frames for Self Forcing and Causal Forcing, 15 latent frames for Rolling Forcing, and 9 latent frames for Reward Forcing in our implementation.

\textbf{Evaluation Metrics.}
Since existing autoregressive video generation models are trained on 5-second video clips, we evaluate \modelname\ and all baselines under two complementary scenarios covering both within-horizon and beyond-horizon generation.
\textit{(1) Long video generation beyond the training horizon.} We assess the ability of \modelname\ to maintain generation quality when video length significantly exceeds the training horizon. We evaluate 30-second and 60-second generation using the VBenchLong~\cite{huang2024vbench} benchmark with prompts from MovieGen~\cite{polyak2024moviegen}.
\textit{(2) Short video generation within the training horizon.} We evaluate KV cache compression quality against baselines under the same Self-Forcing backbone, where the only variable between methods is the KV cache strategy. Since the full KV cache incurs minimal information loss within the training horizon, we treat full-cache generation as the reference and report PSNR~\cite{wang2004ssim}, LPIPS~\cite{zhang2018lpips}, and SSIM~\cite{wang2004ssim} to measure reconstruction fidelity, alongside VBench~\cite{huang2024vbench} scores for general video quality, using prompts from MovieGen~\cite{polyak2024moviegen}. 
All baselines use their officially recommended KV budgets to ensure a fair comparison.

\subsection{Generation Quality Study}

\begin{table*}[t]
\centering
\caption{Quantitative comparison of \modelname\ against baselines for \textbf{5-second short video generation} under reduced KV cache budgets. ``CR'' denotes the KV cache compression ratio relative to full cache. $\uparrow$ indicates higher is better, while $\downarrow$ indicates lower is better. Best results are bold. The results show that \modelname\ incurs less information loss under the reduced KV cache budgets.}
\label{tab:short-video}
\setlength{\tabcolsep}{2pt}
\renewcommand{\arraystretch}{1.15}
\resizebox{\textwidth}{!}{%
\begin{tabular}{c|c|l|ccc|cccc}
\toprule
\textbf{Backbone} & \textbf{Config} & \multicolumn{1}{c|}{\textbf{KV Cache Policy}}
  & \textbf{PSNR}$\uparrow$
  & \textbf{SSIM}$\uparrow$
  & \textbf{LPIPS}$\downarrow$
  & \makecell{\textbf{Subject}\\\textbf{Consist.}$\uparrow$}
  & \makecell{\textbf{Background}\\\textbf{Consist.}$\uparrow$}
  & \makecell{\textbf{Aesthetic}\\\textbf{Quality}$\uparrow$}
  & \makecell{\textbf{Imaging}\\\textbf{Quality}$\uparrow$} \\
\midrule
\multirow{9}{*}{Self Forcing}
  & \multicolumn{2}{c|}{Full Cache} & {-} & {-} & {-} & 94.18 & 94.72 & 63.71 & 69.97 \\
\cmidrule(l){2-10}
  & \multirow{4}{*}{\makecell{KV CR\\=28.57\%}}
    & Reward Forcing  & 18.438 & 0.665 & 0.174 & 94.29 & 94.76 & 63.69 & 70.01 \\
  & & Deep Forcing    & 18.562 & 0.660 & 0.176 & 94.09 & 94.64 & 63.65 & 69.92 \\
  & & Dummy Forcing   & 17.939 & 0.633 & 0.200 & 94.17 & \textbf{94.81} & 63.68 & 70.02 \\
  & & \gmethod{\modelname} & \gb{19.258} & \gb{0.690} & \gb{0.158} & \gb{94.38} & \gc{94.69} & \gb{63.75} & \gb{70.14} \\
\cmidrule(l){2-10}
  & \multirow{4}{*}{\makecell{KV CR\\=42.85\%}}
    & Reward Forcing  & 21.344 & 0.758 & 0.116 & 94.21 & 94.74 & 63.72 & 70.03 \\
  & & Deep Forcing    & 20.943 & 0.744 & 0.122 & 94.14 & 94.69 & 63.67 & 69.90 \\
  & & Dummy Forcing   & 19.333 & 0.683 & 0.161 & \textbf{94.33} & 94.75 & 63.69 & 70.03 \\
  & & \gmethod{\modelname} & \gb{22.164} & \gb{0.780} & \gb{0.106} & \gb{94.33} & \gb{94.77} & \gb{63.77} & \gb{70.09} \\
\midrule
\multirow{9}{*}{Causal Forcing}
  & \multicolumn{2}{c|}{Full Cache} & {-} & {-} & {-} & 94.33 & 94.82 & 64.50 & 70.85 \\
\cmidrule(l){2-10}
  & \multirow{4}{*}{\makecell{KV CR\\=28.57\%}}
    & Reward Forcing  & 16.290 & 0.664 & 0.199 & 94.24 & \textbf{94.82} & 64.39 & \textbf{70.87} \\
  & & Deep Forcing    & 16.047 & 0.651 & 0.206 & 94.17 & 94.65 & 64.23 & 70.63 \\
  & & Dummy Forcing   & 15.191 & 0.612 & 0.249 & 94.18 & 94.66 & 64.15 & 70.64 \\
  & & \gmethod{\modelname} & \gb{16.704} & \gb{0.681} & \gb{0.189} & \gb{94.36} & \gc{94.75} & \gb{64.48} & \gc{70.85} \\
\cmidrule(l){2-10}
  & \multirow{4}{*}{\makecell{KV CR\\=42.85\%}}
    & Reward Forcing  & 18.958 & 0.756 & 0.129 & 94.30 & 94.71 & 64.38 & \textbf{70.88} \\
  & & Deep Forcing    & 18.417 & 0.729 & 0.139 & 94.29 & 94.75 & 64.37 & 70.77 \\
  & & Dummy Forcing   & 16.571 & 0.659 & 0.199 & 94.31 & 94.76 & 64.39 & 70.74 \\
  & & \gmethod{\modelname} & \gb{19.127} & \gb{0.758} & \gb{0.126} & \gb{94.32} & \gb{94.81} & \gb{64.49} & \gc{70.85} \\
\midrule
\multirow{9}{*}{LongLive}
  & \multicolumn{2}{c|}{Full Cache} & {-} & {-} & {-} & 95.85 & 95.92 & 65.12 & 70.09 \\
\cmidrule(l){2-10}
  & \multirow{4}{*}{\makecell{KV CR\\=28.57\%}}
    & Reward Forcing  & 20.001 & 0.734 & 0.126 & 95.97 & 95.97 & 65.10 & 70.08 \\
  & & Deep Forcing    & 15.932 & 0.595 & 0.270 & \textbf{96.56} & \textbf{96.18} & 64.98 & 69.87 \\
  & & Dummy Forcing   & 18.572 & 0.689 & 0.167 & 95.98 & 95.95 & 65.01 & 70.01 \\
  & & \gmethod{\modelname} & \gb{20.830} & \gb{0.761} & \gb{0.109} & \gc{95.90} & \gc{95.97} & \gb{65.12} & \gb{70.15} \\
\cmidrule(l){2-10}
  & \multirow{4}{*}{\makecell{KV CR\\=42.85\%}}
    & Reward Forcing  & 23.313 & 0.824 & 0.075 & \textbf{95.88} & 95.94 & \textbf{65.16} & 70.11 \\
  & & Deep Forcing    & 21.480 & 0.775 & 0.101 & 95.84 & 95.89 & 65.02 & 70.03 \\
  & & Dummy Forcing   & 20.298 & 0.739 & 0.127 & 95.85 & 95.92 & 65.08 & 70.11 \\
  & & \gmethod{\modelname} & \gb{24.035} & \gb{0.842} & \gb{0.065} & \gb{95.88} & \gb{95.95} & \gc{65.11} & \gb{70.14} \\
\bottomrule
\end{tabular}%
}
\end{table*}

\textbf{Long Video Generation (Beyond Training Horizon).}
We evaluate \modelname\ on 30s and 60s long video generation with VBench-Long metrics~\cite{huang2024vbench} using MovieGen prompts~\cite{polyak2024moviegen}, with results in Table~\ref{tab:long-video}. As shown, our \modelname\ outperforms the vanilla backbone and training-free Deep Forcing baseline across all four forcing strategies on most metrics under both configurations, improving overall Subject Consistency and Aesthetic Quality, reflecting better visual fidelity under a reduced cache budget without sacrificing quality. These results demonstrate the superior performance and scalability of \modelname\ for long video generation.

\textbf{Discussion on Dynamic Degree Decrease.}
For the Causal Forcing backbone, \modelname\ exhibits a lower dynamic degree in Table~\ref{tab:long-video}. This does \textbf{not indicate weaker motion}, since higher baseline scores often come from \textbf{abrupt, undesired scene changes} rather than meaningful object or camera motion. By avoiding unstable transitions, \modelname\ preserves \textbf{reasonable dynamics} with better temporal consistency, and more intuitive visual comparisons are provided on our \href{https://anonymous.4open.science/w/FutureForcing}{anonymous website}.

\textbf{Short Video Generation (Within Training Horizon).}
We further evaluate \modelname\ on 5s short video generation within the training horizon, with results in Table~\ref{tab:short-video}. Under a reduced KV cache budget, \modelname\ achieves the closest performance to the full-cache reference among baselines, showing that our future-aware compression incurs minimal information loss. This suggests that as training supports longer horizons, our \modelname\ can efficiently reduce KV cache memory while providing a practical, scalable compression solution for the future long-video generation models.

\subsection{Ablation Study}
\begin{figure}[t]
    \centering
    \begin{minipage}[t]{0.64\linewidth}
        \centering
        \includegraphics[width=\linewidth]{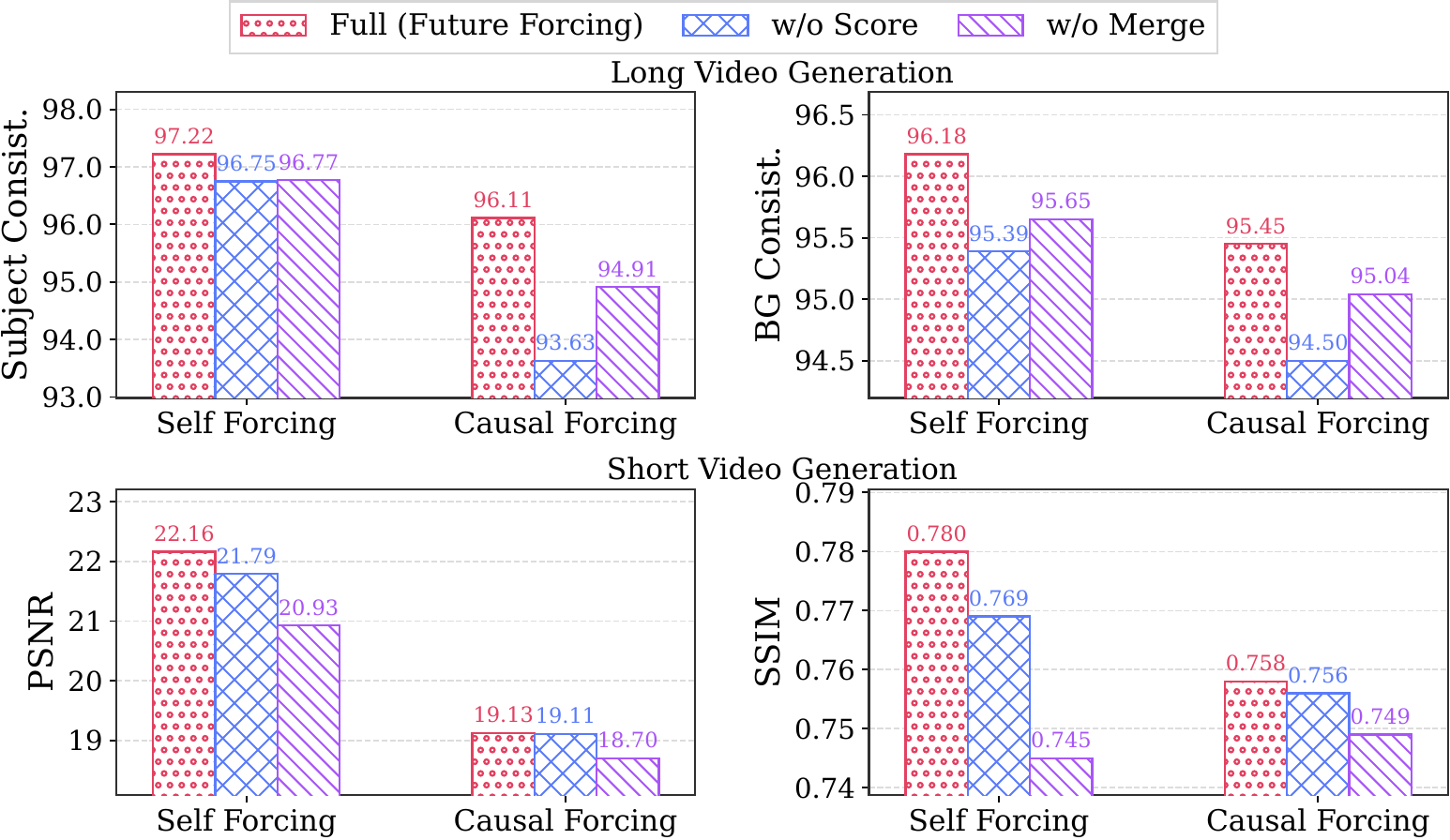}
        \caption{Ablation study of \modelname\ on long- and short-video generation, isolating future scoring and merging modules.}
        \label{fig:ablation}
    \end{minipage}%
    \hspace{0.03\linewidth}%
    \begin{minipage}[t]{0.32\linewidth}
        \centering
        \includegraphics[width=\linewidth]{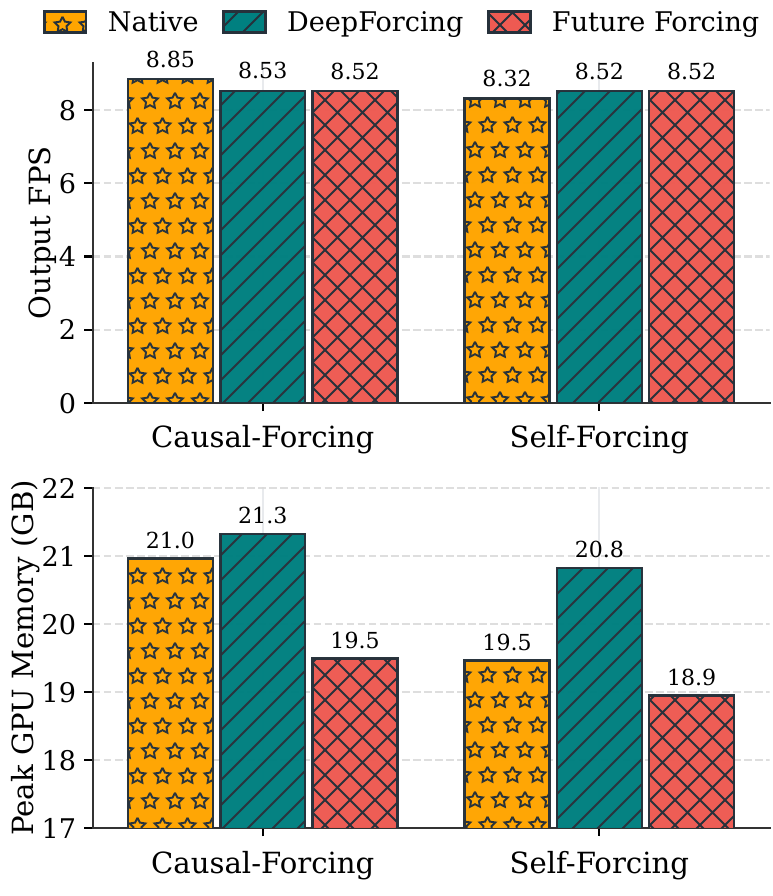}
        \caption{Efficiency and memory study of our \modelname.}
        \label{fig:fps_memory}
    \end{minipage}
    
\end{figure}

To assess each component, we conduct ablations with two variants: (1) \textit{w/o Score}, replacing Future-aware KV Cache Eviction with Sink + FIFO, and (2) \textit{w/o Merge}, removing Future-aware KV Cache Merging and discarding evicted entries. We evaluate them on 30s long-video and 5s short-video generation, respectively. As shown in Figure~\ref{fig:ablation}, both modules improve performance: future-aware eviction contributes more to long-video generation, while merging is critical for short-video fidelity.

\begin{figure}[t]
    \centering
    \vspace{-1em}
    \includegraphics[width=\linewidth]{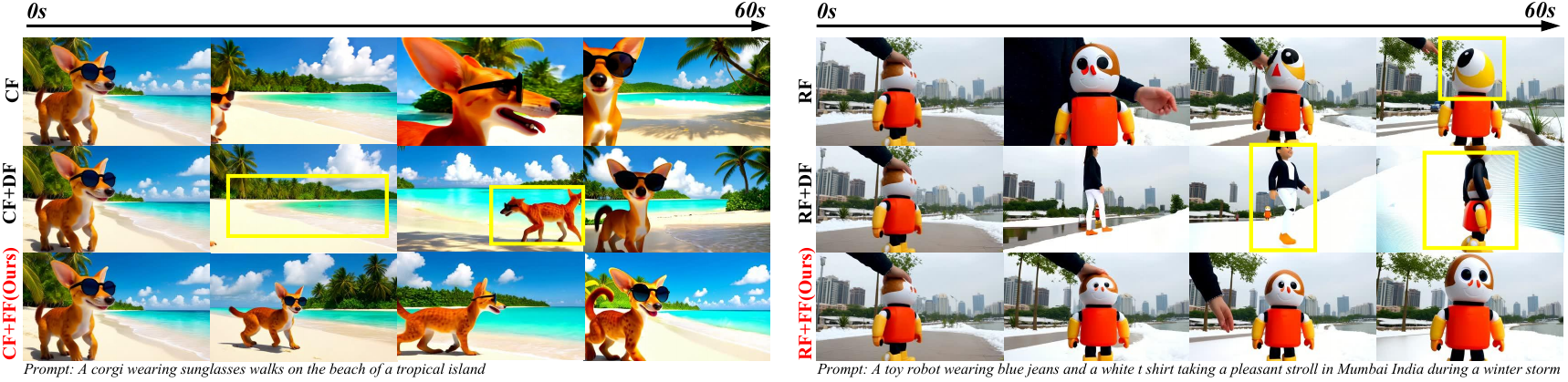}
    \caption{Visual comparison of \modelname\ and baselines for 60-second long-video generation.}
    \vspace{-1em}
    \label{fig:visualization}
\end{figure}

\subsection{Efficiency and Peak GPU Memory Usage Study}
We evaluate the computational efficiency and peak GPU memory of \modelname\ against other KV cache strategies on NVIDIA A100 80GB GPUs, with results reported in Figure~\ref{fig:fps_memory}. The results show that \modelname\ achieves comparable efficiency to existing baselines across both Self Forcing and Causal Forcing while reducing peak GPU memory usage by up to 1.8GB under the Causal Forcing.

\subsection{Visualization}
We visualize generated videos of \modelname\ in Figure~\ref{fig:visualization}. As shown, our \modelname\ produces better temporal consistency across long video generation compared to baselines, as the future-aware KV cache compression alleviates the short-sighted attention problem over long-horizon decoding.\\
\textbf{Additional Results.}
Further results, including hyperparameter analysis, are provided in Appendix~\ref{sec:appendix_exp}.

\section{Conclusion}
In this work, we presented \modelname, a future-aware training-free KV cache policy for AR video generation. We identify a key property of trained AR video models: although attention-used RoPE-modulated queries vary substantially across AR steps, their underlying pre-RoPE query distributions remain stable under a fixed prompt. Based on this insight, \modelname\ constructs future query proxies to guide cache eviction and merges evicted entries in a future-query profiling space to reduce information loss. Extensive experiments across representative AR video generation models show that our proposed \modelname\ improves the long-horizon consistency under limited KV cache budgets.
\newpage

\bibliographystyle{plainnat}
\bibliography{references}

\newpage
\appendix
\renewcommand{\thesection}{\Alph{section}}
\numberwithin{equation}{section}
\numberwithin{theorem}{section}
\setcounter{theorem}{0}
\numberwithin{analysis}{section}
\setcounter{analysis}{0}

\section{Proof}
\label{sec:appendix_proof}

In this section, we prove the proposed Theoretical Analysis~\ref{thorem:upper_bound} in Section~\ref{sec:method}.
We first restate the Theoretical Analysis.

\begin{analysis}[Query Drift Bound]
Given a fixed prompt $c$, for existing trained forcing-based AR video generation transformer $f_{\boldsymbol{\theta}}$, let $\mathbf{X}^t \in \mathbb{R}^{N \times d}$ denote the latent representation at step $t$, and let $\mathbf{Q}^t = \mathbf{X}^t \mathbf{W}_Q^\top$ and $\tilde{\mathbf{Q}}^t$ denote the pre-RoPE and RoPE-modulated query matrices, respectively. We measure query drift via the relative change $\rho_{\mathrm{pre}}^{t \to s} = \|\mathbf{Q}^s - \mathbf{Q}^t\|_F / \|\mathbf{Q}^t\|_F$ and $\rho_{\mathrm{post}}^{t \to s} = \|\tilde{\mathbf{Q}}^s - \tilde{\mathbf{Q}}^t\|_F / \|\tilde{\mathbf{Q}}^t\|_F$. Let $\epsilon = \|\mathbf{X}^s - \mathbf{X}^t\|_F$ denote the latent difference between steps, $\Delta p = p_t - p_s$ the relative positional offset, and $\theta$ the RoPE base frequency. The following two upper bounds then hold:
\begin{equation}
\rho_{\mathrm{pre}}^{t\to s} \leq \frac{\|\mathbf{W}_Q\|_{\mathrm{op}} \cdot \epsilon}{\|\mathbf{Q}^t\|_F}, \qquad
\rho_{\mathrm{post}}^{t\to s} \leq 2\left|\sin\frac{\Delta p \cdot \theta}{2}\right| + \frac{\|\mathbf{W}_Q\|_{\mathrm{op}} \cdot \epsilon}{\|\mathbf{Q}^t\|_F}.
\end{equation}
\end{analysis}

\begin{proof}
We first prove the Pre-RoPE upper bound, then we prove the RoPE-modulated upper bound.

\textbf{(1) Pre-RoPE Bound.}
By definition, $\mathbf{Q}^t=\mathbf{X}^t\mathbf{W}_Q^\top$ and
$\mathbf{Q}^s=\mathbf{X}^s\mathbf{W}_Q^\top$. Therefore,
\begin{equation}
\mathbf{Q}^s-\mathbf{Q}^t
=
(\mathbf{X}^s-\mathbf{X}^t)\mathbf{W}_Q^\top .
\end{equation}
Using the mixed Frobenius--operator norm inequality
$\|\mathbf{A}\mathbf{B}\|_F\leq\|\mathbf{A}\|_F\|\mathbf{B}\|_{\mathrm{op}}$, we have
\begin{equation}
\|\mathbf{Q}^s-\mathbf{Q}^t\|_F
\leq
\|\mathbf{X}^s-\mathbf{X}^t\|_F
\|\mathbf{W}_Q^\top\|_{\mathrm{op}}
=
\epsilon\|\mathbf{W}_Q\|_{\mathrm{op}} .
\end{equation}
Dividing both sides by $\|\mathbf{Q}^t\|_F>0$ gives
\begin{equation}
\rho_{\mathrm{pre}}^{t\to s}
=
\frac{\|\mathbf{Q}^s-\mathbf{Q}^t\|_F}{\|\mathbf{Q}^t\|_F}
\leq
\frac{\epsilon\|\mathbf{W}_Q\|_{\mathrm{op}}}{\|\mathbf{Q}^t\|_F}.
\end{equation}

\textbf{(2) RoPE-Modulated Bound.}
In 3D RoPE, each token $n$ in frame $t$ has position $(p_t,h_n,w_n)$, where
$p_t$ is the temporal position and $(h_n,w_n)$ are spatial positions. Let
$R(p_t,h_n,w_n)$ denote the corresponding block-diagonal RoPE rotation. Since
each $2\times2$ block is orthogonal, $R(p_t,h_n,w_n)$ is orthogonal and
preserves vector norms. Hence,
\begin{equation}
\|\tilde{\mathbf{Q}}^t\|_F
=
\|\mathbf{Q}^t\|_F .
\end{equation}

For token $n$, the RoPE-modulated query difference is
\begin{equation}
\tilde{\mathbf{q}}_s^{(n)}-\tilde{\mathbf{q}}_t^{(n)}
=
R(p_s,h_n,w_n)\mathbf{q}_s^{(n)}
-
R(p_t,h_n,w_n)\mathbf{q}_t^{(n)} .
\end{equation}
Adding and subtracting $R(p_s,h_n,w_n)\mathbf{q}_t^{(n)}$ yields
\begin{equation}
\tilde{\mathbf{q}}_s^{(n)}-\tilde{\mathbf{q}}_t^{(n)}
=
\underbrace{
R(p_s,h_n,w_n)
\bigl(\mathbf{q}_s^{(n)}-\mathbf{q}_t^{(n)}\bigr)
}_{\mathbf{b}^{(n)}}
+
\underbrace{
\bigl(R(p_s,h_n,w_n)-R(p_t,h_n,w_n)\bigr)
\mathbf{q}_t^{(n)}
}_{\mathbf{a}^{(n)}} .
\end{equation}
Let $\mathbf{A}$ and $\mathbf{B}$ collect the rows
$\{\mathbf{a}^{(n)}\}_{n=1}^N$ and $\{\mathbf{b}^{(n)}\}_{n=1}^N$, respectively.
Then
\begin{equation}
\tilde{\mathbf{Q}}^s-\tilde{\mathbf{Q}}^t
=
\mathbf{A}+\mathbf{B}.
\end{equation}
By the triangle inequality,
\begin{equation}
\|\tilde{\mathbf{Q}}^s-\tilde{\mathbf{Q}}^t\|_F
\leq
\|\mathbf{A}\|_F+\|\mathbf{B}\|_F .
\end{equation}

Since RoPE rotations are orthogonal, the content term satisfies
\begin{equation}
\|\mathbf{B}\|_F
=
\|\mathbf{Q}^s-\mathbf{Q}^t\|_F
\leq
\epsilon\|\mathbf{W}_Q\|_{\mathrm{op}} .
\end{equation}

It remains to bound the positional term $\|\mathbf{A}\|_F$. Since token $n$
keeps the same spatial position $(h_n,w_n)$ in frames $t$ and $s$, the height
and width RoPE blocks cancel, leaving only the temporal blocks:
\begin{equation}
\mathbf{a}^{(n)}
=
\bigl(R_{\mathcal{T}}(p_s)-R_{\mathcal{T}}(p_t)\bigr)
\mathbf{q}_{t,\mathcal{T}}^{(n)} ,
\end{equation}
where $R_{\mathcal{T}}(p)$ denotes the temporal RoPE block. Thus,
\begin{equation}
\|\mathbf{a}^{(n)}\|
\leq
\|R_{\mathcal{T}}(p_s)-R_{\mathcal{T}}(p_t)\|_{\mathrm{op}}
\|\mathbf{q}_{t,\mathcal{T}}^{(n)}\| .
\end{equation}
Summing over all tokens and using
$\|\mathbf{q}_{t,\mathcal{T}}^{(n)}\|\leq\|\mathbf{q}_t^{(n)}\|$ gives
\begin{equation}
\|\mathbf{A}\|_F
\leq
\|R_{\mathcal{T}}(p_s)-R_{\mathcal{T}}(p_t)\|_{\mathrm{op}}
\|\mathbf{Q}^t\|_F .
\end{equation}

For each temporal RoPE frequency $\theta_k$, the corresponding rotation block satisfies
\begin{equation}
\|R_2(p_s\theta_k)-R_2(p_t\theta_k)\|_{\mathrm{op}}
=
2\left|\sin\frac{(p_s-p_t)\theta_k}{2}\right| .
\end{equation}
Let $\Delta p=p_t-p_s$ and $\theta$ be the temporal RoPE frequency. Then
\begin{equation}
\|R_{\mathcal{T}}(p_s)-R_{\mathcal{T}}(p_t)\|_{\mathrm{op}}
=
2\left|\sin\frac{\Delta p\,\theta}{2}\right| .
\end{equation}
Therefore,
\begin{equation}
\|\mathbf{A}\|_F
\leq
2\left|\sin\frac{\Delta p\,\theta}{2}\right|
\|\mathbf{Q}^t\|_F .
\end{equation}

Combining the bounds for $\mathbf{A}$ and $\mathbf{B}$, we obtain
\begin{equation}
\|\tilde{\mathbf{Q}}^s-\tilde{\mathbf{Q}}^t\|_F
\leq
2\left|\sin\frac{\Delta p\,\theta}{2}\right|
\|\mathbf{Q}^t\|_F
+
\epsilon\|\mathbf{W}_Q\|_{\mathrm{op}} .
\end{equation}
Finally, using $\|\tilde{\mathbf{Q}}^t\|_F=\|\mathbf{Q}^t\|_F$ and dividing by
$\|\tilde{\mathbf{Q}}^t\|_F>0$ gives
\begin{equation}
\rho_{\mathrm{post}}^{t\to s}
=
\frac{\|\tilde{\mathbf{Q}}^s-\tilde{\mathbf{Q}}^t\|_F}
{\|\tilde{\mathbf{Q}}^t\|_F}
\leq
2\left|\sin\frac{\Delta p\,\theta}{2}\right|
+
\frac{\epsilon\|\mathbf{W}_Q\|_{\mathrm{op}}}{\|\mathbf{Q}^t\|_F}.
\end{equation}
This completes the proof.
\end{proof}

\section{Custom Kernel Design}
\label{sec:appendix_kernel}

Computing the per-token importance scores in Eq.~\eqref{eq:score} requires reducing the softmax attention weights $\mathrm{softmax}(\mathbf{Q}\mathbf{K}^{\top}/\sqrt{d_h})$ over the query dimension. Standard attention kernels, such as FlashAttention~\citep{dao2022flashattention,dao2023flashattention2}, do not expose these weights, since they are designed to return only the value aggregated attention output while avoiding materialization of the intermediate softmax matrix. A naive dense PyTorch implementation must therefore instantiate the full $\mathbf{Q}\mathbf{K}^{\top}$ probability matrix before reduction. To address this, we implement a custom two-pass Triton kernel that evaluates Eq.~\eqref{eq:score} without instantiating the full attention weight matrix, adapting the online softmax primitive from value aggregation to score reduction.

\paragraph{Algorithm.}
The kernel specializes the online softmax recurrence for the reduction required by Eq.~\eqref{eq:score}. Instead of accumulating a weighted sum of values, it accumulates the softmax mass assigned to each key slot. This requires two passes over $\mathbf{K}$.

\textbf{Pass 1} streams $\mathbf{K}$ in tiles of size $B_K$, while maintaining the per query running maximum $m_q$ and partial normalization sum $\ell_q$ in SRAM. After all key tiles have been processed, the kernel writes the finalized log sum exp value
\begin{equation}
    \mathrm{LSE}_{b,h,q} = m_q + \log \ell_q
\end{equation}
to an $O(BHQ)$ buffer. This pass is necessary because normalized attention weights cannot be computed until the softmax denominator has been finalized.

\textbf{Pass 2} streams $\mathbf{K}$ again using the same $(BH,\,\lceil Q/B_Q\rceil)$ launch grid. For each tile, the kernel recomputes the raw dot product
\begin{equation}
    s_{q,t} = \mathbf{Q}_q\mathbf{K}_t^{\top}/\sqrt{d_h}
\end{equation}
on chip, applies the finalized normalizer,
\begin{equation}
    w_{q,t} = \exp\!\left(s_{q,t} - \mathrm{LSE}_{b,h,q}\right),
\end{equation}
and immediately reduces over the query tile to accumulate $\sum_{q \in \mathrm{tile}} w_{q,t}$ for each key slot. Summing these partial contributions across query tiles yields $\mathrm{score}_{b,h,t}$ as defined in Eq.~\eqref{eq:score}.

\paragraph{Implementation Details.}
Both passes use tile sizes $B_Q{=}B_K{=}128$ and share the same launch configuration. All tile level intermediates, including $\mathbf{Q}$ tiles, $\mathbf{K}$ tiles, running statistics $m_q$ and $\ell_q$, and normalized weights $w_{q,t}$, are kept in SRAM during each tile iteration. The head dimension $d_h$ is padded to the next power of two to support valid \texttt{tl.dot} operations for arbitrary head widths. The resulting working set is
\begin{equation}
    O\!\left(BHQ + BH\left\lceil \frac{Q}{B_Q} \right\rceil T\right),
\end{equation}
which avoids the dense $O(BHQT)$ softmax matrix and makes per-token score computation practical for long video generation.

\begin{figure}[t]
\centering
\begin{subfigure}[t]{0.495\textwidth}
    \centering
    \includegraphics[width=\linewidth]{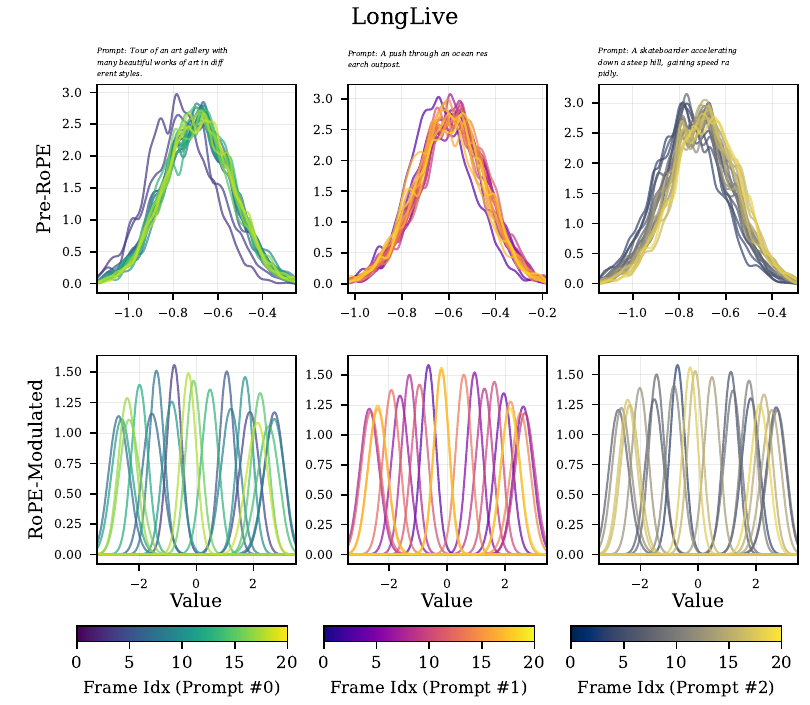}
    \caption{Dimension = 0}
    \label{fig:longlive_query_dim0}
\end{subfigure}%
\hfill%
\begin{subfigure}[t]{0.495\textwidth}
    \centering
    \includegraphics[width=\linewidth]{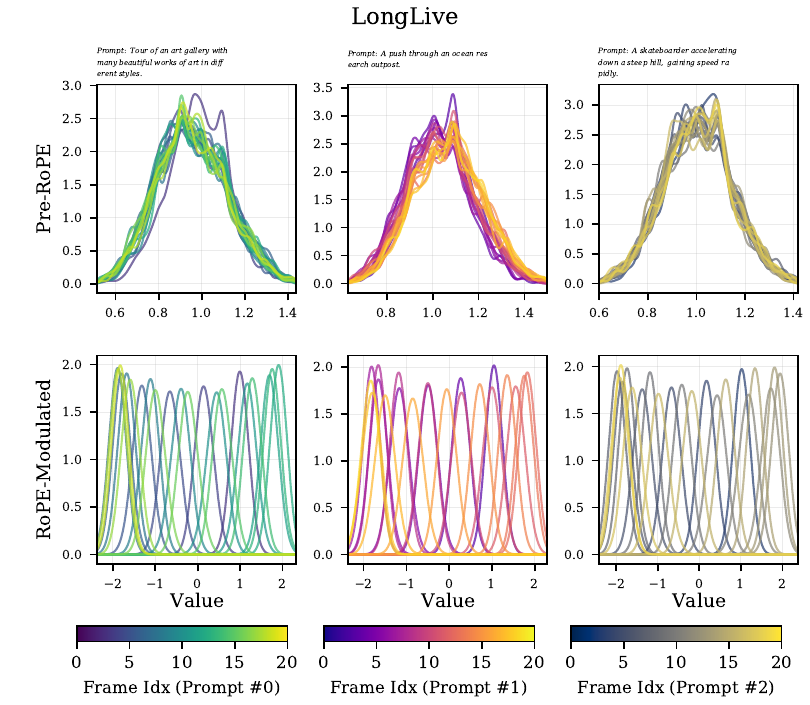}
    \caption{Dimension = 5}
    \label{fig:longlive_query_dim5}
\end{subfigure}
\caption{
Query distribution visualizations on LongLive across two representative query dimensions.
}
\label{fig:longlive_query_distribution}
\end{figure}

\begin{figure}[t]
\centering
\begin{subfigure}[t]{0.495\textwidth}
    \centering
    \includegraphics[width=\linewidth]{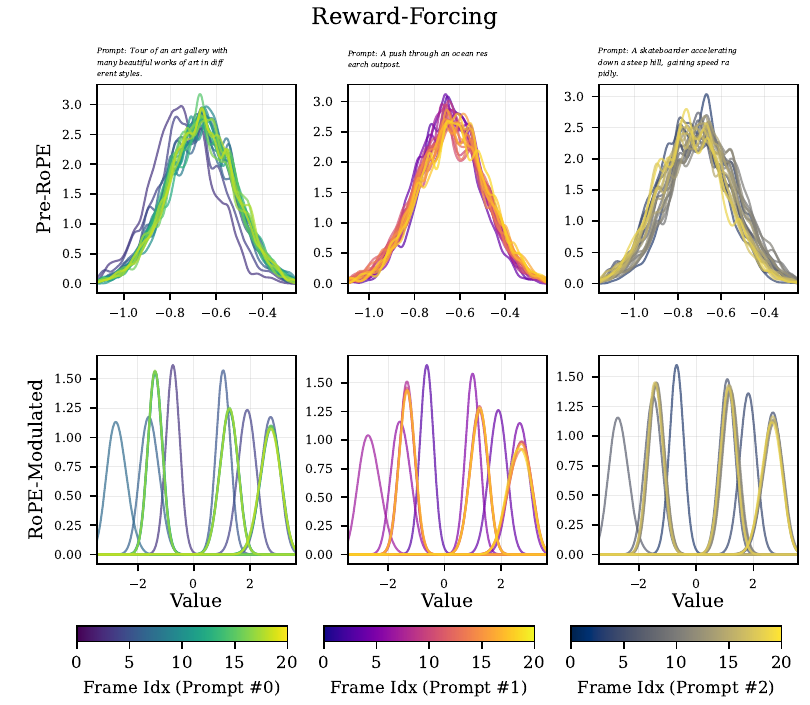}
    \caption{Dimension = 0}
    \label{fig:rewardforcing_query_dim0}
\end{subfigure}%
\hfill%
\begin{subfigure}[t]{0.495\textwidth}
    \centering
    \includegraphics[width=\linewidth]{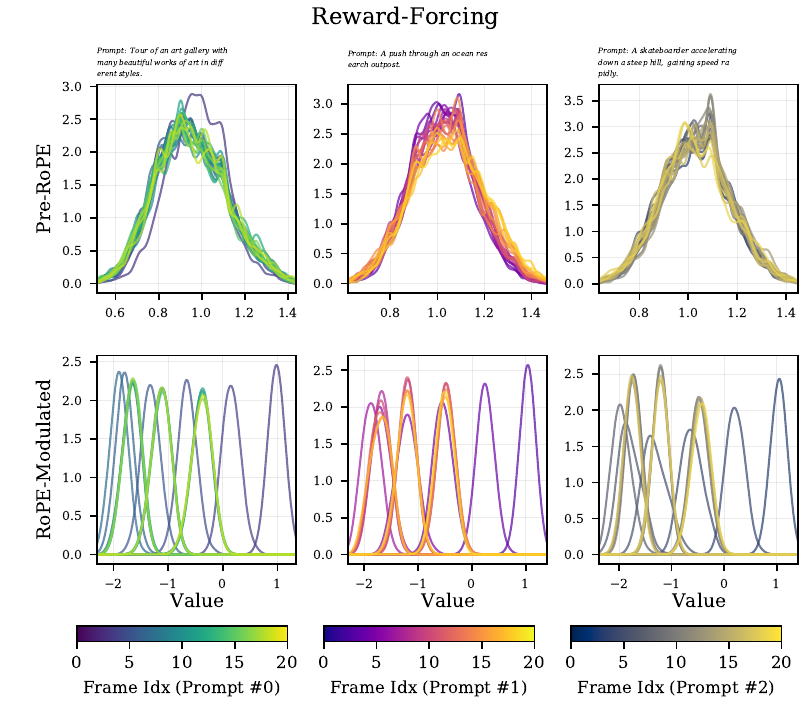}
    \caption{Dimension = 5}
    \label{fig:rewardforcing_query_dim5}
\end{subfigure}
\caption{
Query distribution visualizations on Reward-Forcing across two representative query dimensions.
}
\label{fig:rewardforcing_query_distribution}
\end{figure}

\begin{figure}[t]
\centering
\begin{subfigure}[t]{0.495\textwidth}
    \centering
    \includegraphics[width=\linewidth]{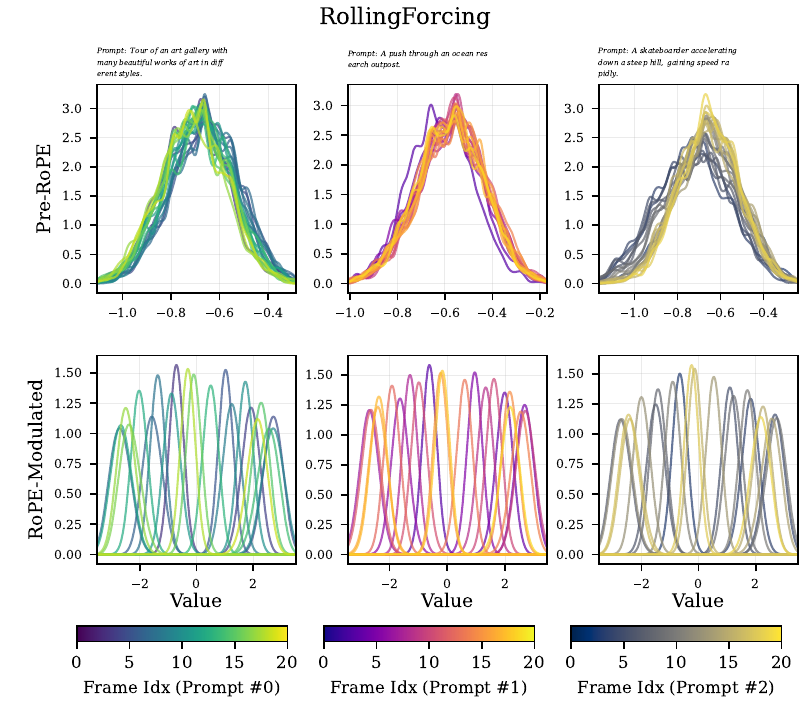}
    \caption{Dimension = 0}
    \label{fig:rollingforcing_query_dim0}
\end{subfigure}%
\hfill%
\begin{subfigure}[t]{0.495\textwidth}
    \centering
    \includegraphics[width=\linewidth]{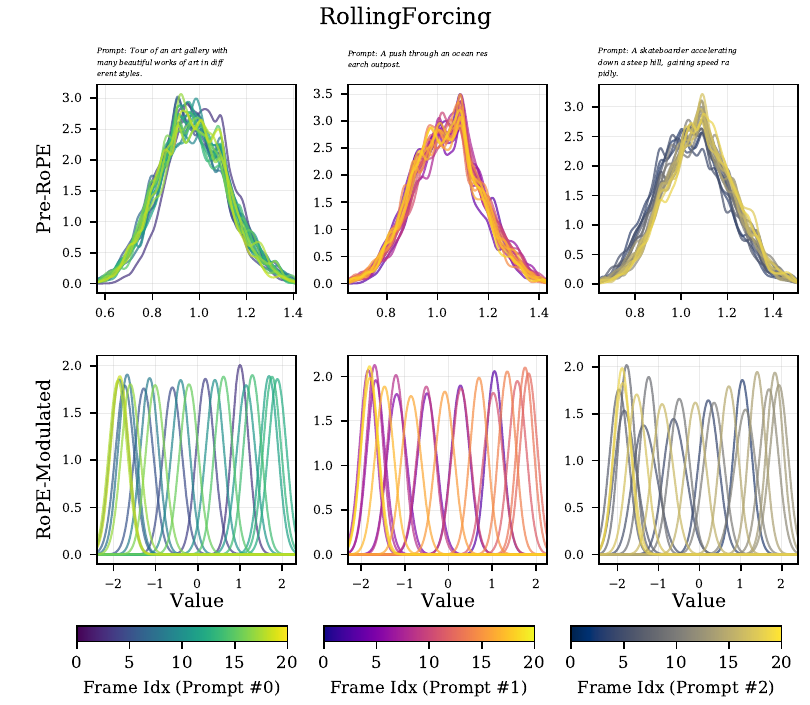}
    \caption{Dimension = 5}
    \label{fig:rollingforcing_query_dim5}
\end{subfigure}
\caption{
Query distribution visualizations on Rolling-Forcing across two representative query dimensions.
}
\label{fig:rollingforcing_query_distribution}
\end{figure}

\section{Additional Experiment Results}
\label{sec:appendix_exp}

\subsection{Additional Query Distribution Analysis Results.}
We provide additional query distribution visualizations to further validate the pre-RoPE query stability discussed in Section~\ref{sec:emprical_analysis}. In particular, we extend the analysis beyond the main-text examples to three additional representative autoregressive video generation models: LongLive~\citep{yang2025longlive}, Reward-Forcing~\citep{lu2025rewardforcing}, and Rolling-Forcing~\citep{liu2025rolling}. For each model, we visualize query distributions at two representative query dimensions, as shown in Figure~\ref{fig:longlive_query_distribution}, Figure~\ref{fig:rewardforcing_query_distribution}, and Figure~\ref{fig:rollingforcing_query_distribution}.
Across all models and dimensions, we observe a consistent pattern: the canonical pre-RoPE query distributions remain concentrated and stable across autoregressive generation steps under a fixed prompt, whereas the corresponding RoPE-modulated query distributions exhibit substantially larger temporal variation. This result is consistent with the observation in Figure~\ref{fig:query_distribution}, and further supports our claim that the dominant temporal drift in attention queries is introduced by RoPE-based positional modulation rather than by changes in the underlying semantic query representation. Therefore, historical pre-RoPE query statistics provide a reliable basis for estimating future attention behavior, which motivates the future-query proxy construction used in our \modelname\ eviction and merging modules.

\begin{figure}[t]
    \centering
    \includegraphics[width=0.8\linewidth]{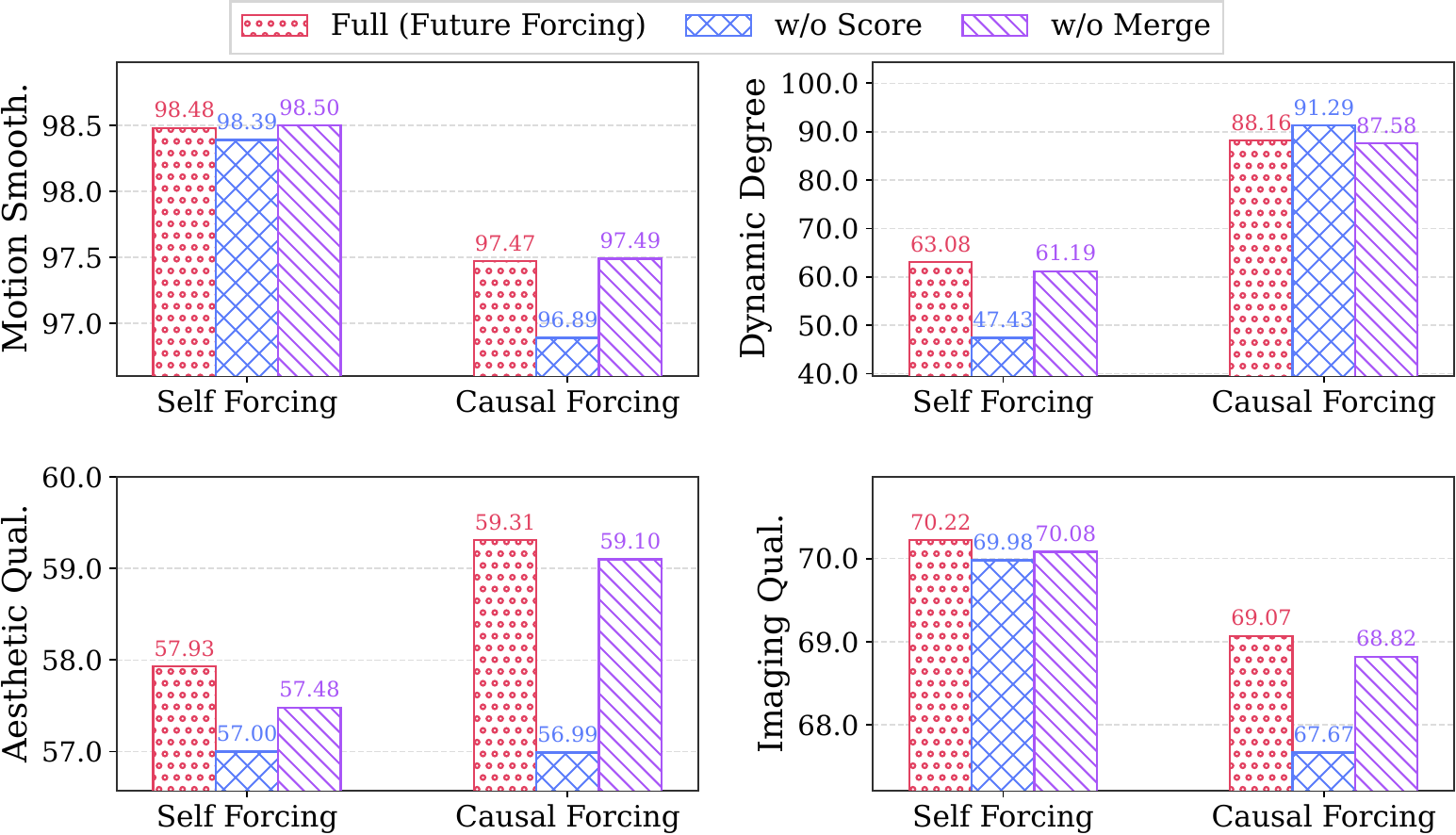}
    \caption{Additional ablation results}
    \label{fig:ablation_appendix}
\end{figure}

\subsection{Additional Ablation Results.}
We further report additional ablation results on VBench dimensions beyond the main text, including Motion Smoothness, Dynamic Degree, Aesthetic Quality, and Imaging Quality. As shown in Figure~\ref{fig:ablation_appendix}, the full \modelname\ maintains competitive or superior overall performance across both Self Forcing and Causal Forcing settings. In particular, removing the future-aware eviction score causes noticeable degradation on Aesthetic Quality and Imaging Quality under Causal Forcing, indicating that standard cache eviction fails to reliably preserve future-relevant tokens when the cache budget is limited. Removing the merging module also degrades several metrics, showing that directly discarding evicted entries introduces non-negligible information loss. Although individual metrics may fluctuate slightly, the full model provides the best overall trade-off across generation smoothness, dynamics, and visual quality. These results further support the necessity of both proposed components: future-aware eviction improves token selection, while future-aware merging reduces attention-output distortion after compression.

\begin{table*}[t]
\centering
\caption{Quantitative comparison across models, KV cache compression ratios, and forcing strategies on additional VBench Motion Smoothness and Dynamic Degree. $\uparrow$ denotes higher is better.}
\label{tab:short-video-appendix}
\setlength{\tabcolsep}{8pt}
\renewcommand{\arraystretch}{1.2}
\resizebox{\textwidth}{!}{%
\begin{tabular}{c|c|l|cc}
\toprule
\textbf{Backbone} & \textbf{Config} & \multicolumn{1}{c|}{\textbf{Method}}
  & \textbf{Motion Smoothness}$\uparrow$
  & \textbf{Dynamic Degree}$\uparrow$ \\
\midrule
\multirow{10}{*}{Self Forcing}
  & \multicolumn{2}{c|}{Full Cache} & 98.32 & 61.91 \\
\cmidrule(l){2-5}
  & \multirow{4}{*}{KV Cache CR=28.57\%}
    & Reward Forcing  & 98.29 & 62.41 \\
  & & Deep Forcing    & 98.30 & 62.17 \\
  & & Dummy Forcing   & 98.37 & 59.52 \\
  & & \gmethod{\modelname} & \gc{98.43} & \gc{61.42} \\
\cmidrule(l){2-5}
  & \multirow{4}{*}{KV Cache CR=42.85\%}
    & Reward Forcing  & 98.32 & 61.71 \\
  & & Deep Forcing    & 98.29 & 62.31 \\
  & & Dummy Forcing   & 98.35 & 60.52 \\
  & & \gmethod{\modelname} & \gc{98.35} & \gc{61.78} \\
\midrule
\multirow{10}{*}{Causal Forcing}
  & \multicolumn{2}{c|}{Full Cache} & 97.67 & 76.87 \\
\cmidrule(l){2-5}
  & \multirow{4}{*}{KV Cache CR=28.57\%}
    & Reward Forcing  & 97.59 & 78.46 \\
  & & Deep Forcing    & 97.55 & 78.46 \\
  & & Dummy Forcing   & 97.59 & 79.86 \\
  & & \gmethod{\modelname} & \gc{97.58} & \gc{79.92} \\
\cmidrule(l){2-5}
  & \multirow{4}{*}{KV Cache CR=42.85\%}
    & Reward Forcing  & 97.62 & 78.27 \\
  & & Deep Forcing    & 97.64 & 77.77 \\
  & & Dummy Forcing   & 97.60 & 78.07 \\
  & & \gmethod{\modelname} & \gc{97.64} & \gc{78.17} \\
\midrule
\multirow{10}{*}{LongLive}
  & \multicolumn{2}{c|}{Full Cache} & 98.94 & 37.19 \\
\cmidrule(l){2-5}
  & \multirow{4}{*}{KV Cache CR=28.57\%}
    & Reward Forcing  & 98.82 & 35.19 \\
  & & Deep Forcing    & 98.80 & 39.48 \\
  & & Dummy Forcing   & 98.75 & 35.59 \\
  & & \gmethod{\modelname} & \gc{98.95} & \gc{37.69} \\
\cmidrule(l){2-5}
  & \multirow{4}{*}{KV Cache CR=42.85\%}
    & Reward Forcing  & 98.94 & 36.39 \\
  & & Deep Forcing    & 98.90 & 37.39 \\
  & & Dummy Forcing   & 98.95 & 35.39 \\
  & & \gmethod{\modelname} & \gc{98.94} & \gc{37.49} \\
\bottomrule
\end{tabular}%
}
\end{table*}

\subsection{Additional Generation Quality Results.}

We provide additional short-video generation results on VBench motion-related dimensions in Table~\ref{tab:short-video-appendix}. These experiments complement the reconstruction-based metrics reported in Table~\ref{tab:short-video} by evaluating whether KV cache compression preserves temporal quality and motion dynamics within the training horizon. All methods are evaluated under the same reduced KV cache budgets, and the full-cache setting is included as a reference upper bound.
Overall, \modelname\ maintains competitive or superior performance on both Motion Smoothness and Dynamic Degree across different backbones and compression ratios. On Self-Forcing, \modelname\ achieves the best Motion Smoothness at the 28.57\% cache ratio and remains comparable to the strongest baselines at the 42.85\% ratio, while preserving similar Dynamic Degree to the full-cache reference. On Causal-Forcing, \modelname\ consistently achieves the best or near-best Dynamic Degree under both cache ratios, indicating that future-aware cache selection can preserve motion intensity without introducing temporal instability. On LongLive, \modelname\ recovers the full-cache Motion Smoothness and improves Dynamic Degree over most compressed baselines, especially under the 28.57\% cache ratio. These results further show that \modelname\ preserves motion quality under constrained KV cache budgets.

\subsection{Additional Efficiency and Memory Consumption Analysis Results.}
\begin{figure}[ht]
    \centering
    \includegraphics[width=0.95\linewidth]{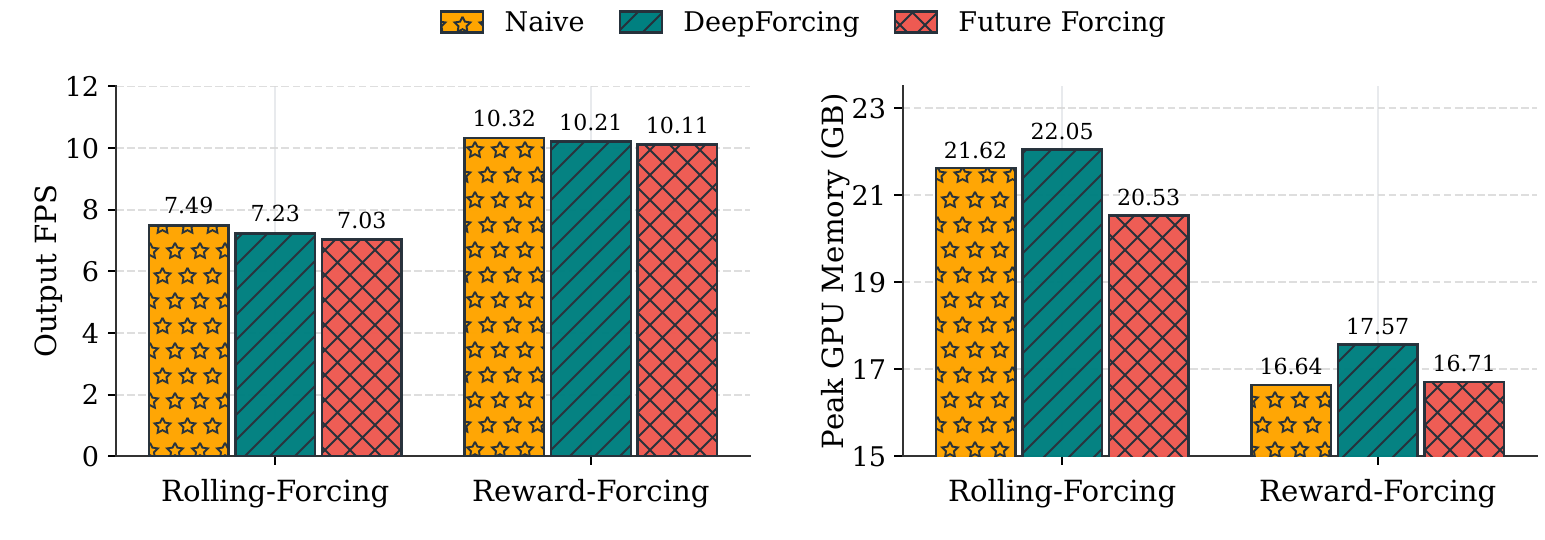}
    \caption{Additional efficiency and memory consumption analysis under different AR video generation backbones.}
    \label{fig:efficiency_memory_appendix}
\end{figure}

We further evaluate the inference efficiency and peak GPU memory consumption of \modelname\ under different forcing strategies, with results shown in Figure~\ref{fig:efficiency_memory_appendix}. Compared with Naive and DeepForcing, \modelname\ achieves comparable output FPS while consistently reducing peak GPU memory usage. Under Rolling-Forcing, \modelname\ attains an output FPS of 7.03, close to Naive and DeepForcing, while reducing peak memory from 21.62GB and 22.05GB to 20.53GB. Under Reward-Forcing, \modelname\ maintains a similar output FPS of 10.11, compared with 10.32 for Naive and 10.21 for DeepForcing, and lowers peak memory from 16.64GB and 17.57GB to 16.71GB, indicating that \modelname\ introduces negligible inference overhead, demonstrating its practical efficiency under constrained KV cache.

\textbf{Reason for lower memory consumption.}
In our experiments, under the Rolling-Forcing setting, \modelname\ uses a smaller KV cache budget than the baselines while achieving better or comparable performance. This further demonstrates the effectiveness of our future-aware KV cache policy: \modelname\ can retain more future-relevant historical tokens under a tighter memory budget, thereby reducing peak GPU memory consumption without sacrificing generation quality.

\subsection{Custom Triton Contribution Analysis Results.}
\begin{figure}[ht]
    \centering
    \includegraphics[width=0.8\linewidth]{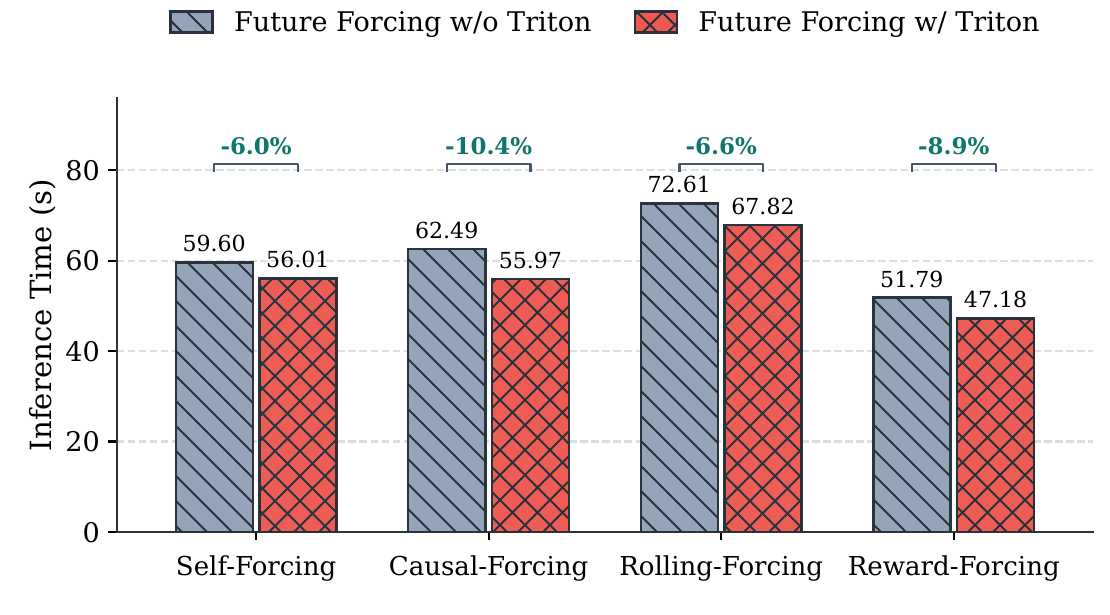}
    \caption{Contribution of the custom Triton kernel to inference efficiency.}
    \label{fig:triton_appendix}
\end{figure}
To evaluate the contribution of our custom Triton kernel, we compare \modelname\ with and without Triton across different forcing strategies, as shown in Figure~\ref{fig:triton_appendix}. The Triton kernel consistently reduces inference time by 6.0\%, 10.4\%, 6.6\%, and 8.9\% under Self-Forcing, Causal-Forcing, Rolling-Forcing, and Reward-Forcing, respectively. This improvement mainly comes from accelerating the token-level importance scoring step, showing that our Triton implementation effectively reduces the computational overhead of future-aware scoring.

\subsection{Additional Hyperparameter Analysis Results.}
To further examine the sensitivity of \modelname\ to its key hyperparameters, we conduct additional analyses on two representative autoregressive video generation backbones, Self-Forcing~\cite{huang2025self} and Causal-Forcing~\cite{zhu2026causalforcing}. We consider two key factors in our future-aware KV cache policy design: the future query proxy horizon $\Delta_{\max}$, which controls the range of future positions used to construct future query proxies, and the future-history balance coefficient $\lambda$, which balances predicted future attention importance and historical attention evidence in the cache eviction score. We evaluate $\Delta_{\max}\in\{1,3,6,9,12\}$ and $\lambda\in\{0.1,0.3,0.5,0.7,0.9\}$, where larger $\lambda$ assigns greater weight to predicted future attention importance. For all experiments, we randomly select 20 prompts from the MovieGen~\cite{polyak2024moviegen} prompt set, keep all other hyperparameters fixed, and use the same initial noise for each prompt across different hyperparameter settings to ensure fair comparison. We save the generated videos for every configuration and evaluate them using VBench-Long~\cite{huang2024vbench} metrics across all evaluation dimensions.

As shown in Table~\ref{tab:self_forcing_delta_max_analysis}, Table~\ref{tab:self_forcing_lambda_analysis}, Table~\ref{tab:causal_forcing_delta_max_analysis} and Table~\ref{tab:causal_forcing_lambda_analysis}, \modelname\ remains robust across a wide range of hyperparameter settings on both backbones. Varying $\Delta_{\max}$ leads to only moderate changes, indicating that the future query proxy construction is not overly sensitive to the exact horizon. For $\lambda$, assigning more weight to future attention importance generally improves motion-related performance, especially on Self-Forcing, where a larger $\lambda$ will increase dynamic degree while only slightly reducing subject and background consistency. On Causal-Forcing, performance remains stable across different $\lambda$ values, with $\lambda=0.7$ providing a strong balance among motion, consistency, and visual quality. Overall, these results show that \modelname\ consistently maintains strong video quality and temporal consistency under fixed KV cache budgets across most hyperparameter configurations.

\begin{table}[t]
\centering
\small
\caption{Hyperparameter analysis of the future query proxy horizon $\Delta_{\max}$ on the Self-Forcing backbone, evaluated on 20 randomly selected prompts from MovieGen~\cite{polyak2024moviegen}.}
\label{tab:self_forcing_delta_max_analysis}
\resizebox{\linewidth}{!}{
\setlength{\tabcolsep}{3pt}
\begin{tabular}{c c c c c c c c}
\toprule
\textbf{Value}
& \makecell[c]{\textbf{Dynamic}\\\textbf{Degree}$\uparrow$}
& \makecell[c]{\textbf{Motion}\\\textbf{Smoothness}$\uparrow$}
& \makecell[c]{\textbf{Overall}\\\textbf{Consistency}$\uparrow$}
& \makecell[c]{\textbf{Imaging}\\\textbf{Quality}$\uparrow$}
& \makecell[c]{\textbf{Aesthetic}\\\textbf{Quality}$\uparrow$}
& \makecell[c]{\textbf{Subject}\\\textbf{Consistency}$\uparrow$}
& \makecell[c]{\textbf{Background}\\\textbf{Consistency}$\uparrow$} \\
\midrule
1  & 56.00 & 98.55 & 25.20 & 72.05 & 55.93 & 97.69 & 96.52 \\
3  & 61.00 & 98.46 & 25.58 & 72.29 & 55.56 & 97.61 & 96.50 \\
6  & 54.33 & 98.55 & 25.42 & 72.19 & 55.67 & 97.69 & 96.52 \\
9  & 56.67 & 98.49 & 25.59 & 71.80 & 56.25 & 97.66 & 96.49 \\
12 & 54.00 & 98.50 & 25.83 & 72.42 & 55.82 & 97.70 & 96.48 \\
\bottomrule
\end{tabular}
}
\end{table}

\begin{table}[t]
\centering
\small
\caption{Hyperparameter analysis of the future-history balance coefficient $\lambda$ on the Self-Forcing backbone, evaluated on 20 randomly selected prompts from MovieGen~\cite{polyak2024moviegen}.}
\label{tab:self_forcing_lambda_analysis}
\resizebox{\linewidth}{!}{
\setlength{\tabcolsep}{3pt}
\begin{tabular}{c c c c c c c c}
\toprule
\textbf{Value}
& \makecell[c]{\textbf{Dynamic}\\\textbf{Degree}$\uparrow$}
& \makecell[c]{\textbf{Motion}\\\textbf{Smoothness}$\uparrow$}
& \makecell[c]{\textbf{Overall}\\\textbf{Consistency}$\uparrow$}
& \makecell[c]{\textbf{Imaging}\\\textbf{Quality}$\uparrow$}
& \makecell[c]{\textbf{Aesthetic}\\\textbf{Quality}$\uparrow$}
& \makecell[c]{\textbf{Subject}\\\textbf{Consistency}$\uparrow$}
& \makecell[c]{\textbf{Background}\\\textbf{Consistency}$\uparrow$} \\
\midrule
0.1 & 53.00 & 98.53 & 25.71 & 72.10 & 55.81 & 97.60 & 96.57 \\
0.3 & 51.67 & 98.53 & 25.42 & 72.32 & 55.76 & 97.70 & 96.57 \\
0.5 & 56.67 & 98.49 & 25.59 & 72.80 & 56.27 & 97.67 & 96.51 \\
0.7 & 62.00 & 98.35 & 25.51 & 72.15 & 55.58 & 97.57 & 96.37 \\
0.9 & 65.67 & 98.25 & 25.23 & 72.25 & 55.99 & 97.28 & 96.20 \\
\bottomrule
\end{tabular}
}
\end{table}

\begin{table}[t]
\centering
\small
\caption{Hyperparameter analysis of the future query proxy horizon $\Delta_{\max}$ on the Causal-Forcing backbone, evaluated on 20 randomly selected prompts from MovieGen~\cite{polyak2024moviegen}.}
\label{tab:causal_forcing_delta_max_analysis}
\resizebox{\linewidth}{!}{
\setlength{\tabcolsep}{3pt}
\begin{tabular}{c c c c c c c c}
\toprule
\textbf{Value}
& \makecell[c]{\textbf{Dynamic}\\\textbf{Degree}$\uparrow$}
& \makecell[c]{\textbf{Motion}\\\textbf{Smoothness}$\uparrow$}
& \makecell[c]{\textbf{Overall}\\\textbf{Consistency}$\uparrow$}
& \makecell[c]{\textbf{Imaging}\\\textbf{Quality}$\uparrow$}
& \makecell[c]{\textbf{Aesthetic}\\\textbf{Quality}$\uparrow$}
& \makecell[c]{\textbf{Subject}\\\textbf{Consistency}$\uparrow$}
& \makecell[c]{\textbf{Background}\\\textbf{Consistency}$\uparrow$} \\
\midrule
1  & 94.33 & 97.13 & 23.84 & 71.35 & 57.53 & 95.93 & 95.39 \\
3  & 90.00 & 97.07 & 24.01 & 71.22 & 57.55 & 96.01 & 95.41 \\
6  & 90.33 & 97.08 & 24.10 & 71.43 & 57.81 & 96.09 & 95.50 \\
9  & 94.33 & 97.23 & 24.20 & 71.18 & 57.70 & 96.05 & 95.56 \\
12 & 90.67 & 97.27 & 24.31 & 71.12 & 57.59 & 95.96 & 95.52 \\
\bottomrule
\end{tabular}
}
\end{table}

\begin{table}[t]
\centering
\small
\caption{Hyperparameter analysis of the future-history balance coefficient $\lambda$ on the Causal-Forcing backbone, evaluated on 20 randomly selected prompts from MovieGen~\cite{polyak2024moviegen}.}
\label{tab:causal_forcing_lambda_analysis}
\resizebox{\linewidth}{!}{
\setlength{\tabcolsep}{3pt}
\begin{tabular}{c c c c c c c c}
\toprule
\textbf{Value}
& \makecell[c]{\textbf{Dynamic}\\\textbf{Degree}$\uparrow$}
& \makecell[c]{\textbf{Motion}\\\textbf{Smoothness}$\uparrow$}
& \makecell[c]{\textbf{Overall}\\\textbf{Consistency}$\uparrow$}
& \makecell[c]{\textbf{Imaging}\\\textbf{Quality}$\uparrow$}
& \makecell[c]{\textbf{Aesthetic}\\\textbf{Quality}$\uparrow$}
& \makecell[c]{\textbf{Subject}\\\textbf{Consistency}$\uparrow$}
& \makecell[c]{\textbf{Background}\\\textbf{Consistency}$\uparrow$} \\
\midrule
0.1 & 89.00 & 97.14 & 24.03 & 71.32 & 58.12 & 96.10 & 95.47 \\
0.3 & 92.00 & 97.11 & 23.66 & 71.28 & 57.60 & 96.12 & 95.43 \\
0.5 & 91.33 & 97.08 & 24.08 & 71.43 & 57.79 & 96.11 & 95.50 \\
0.7 & 92.33 & 97.25 & 23.98 & 71.71 & 57.70 & 96.17 & 95.52 \\
0.9 & 89.33 & 97.23 & 23.92 & 71.44 & 57.90 & 96.11 & 95.54 \\
\bottomrule
\end{tabular}
}
\end{table}

\subsection{Difference to PaFu-KV}
PaFu-KV~\cite{chen2026past} and our \modelname\ both aim to improve KV cache management for autoregressive video generation by incorporating future-aware token importance. However, the two methods differ fundamentally in how future relevance is obtained and used. \textbf{PaFu-KV is a training-based} method that relies on a bidirectional teacher model to provide past- and future-informed salience supervision, and further trains a lightweight salience estimation head to predict token-level importance during autoregressive inference. In contrast, \textbf{\modelname\ is fully training-free}: it does not require an additional teacher model, auxiliary salience head, or any extra fine-tuning. Instead, \modelname\ exploits the empirical stability of canonical pre-RoPE query distributions across autoregressive steps to construct future query proxies, which are then used to estimate future attention importance directly within the pretrained AR video generator. Moreover, while PaFu-KV primarily performs salience-based cache eviction, \modelname\ further introduces future-aware KV cache merging, which preserves information from evicted entries by merging them according to their predicted future attention profiles. This design reduces the attention-output distortion caused by token removal and enables future-aware KV cache compression without retraining the model. \textbf{Since the source code of PaFu-KV is not publicly available, we do not include it in direct comparisons.}

\section{Experiments Compute Resources.}
All experiments are conducted on NVIDIA A100 GPUs with 80GB memory. Since our KV cache policy \modelname\ is training-free, the computational cost mainly comes from autoregressive inference and KV-cache scoring and merging during the autoregressive video generation process.

\section{Limitations}
\label{sec:limitations_appendix}
Although our proposed KV cache policy \modelname\ is training-free and effective across representative autoregressive video generation models, our current study still has several limitations. First, our evaluation mainly focuses on autoregressive video generation, and extending the same future-aware compression principle to other autoregressive multimodal generation tasks remains an interesting direction for future work. Second, our experiments are conducted on widely used open-sourced research backbones and standard evaluation settings, while larger-scale evaluations on more diverse prompts, resolutions, and deployment scenarios could further strengthen the empirical validation.

\section{Additional Visualization Results.}
\begin{figure}[ht]
    \centering
    \includegraphics[width=\linewidth]{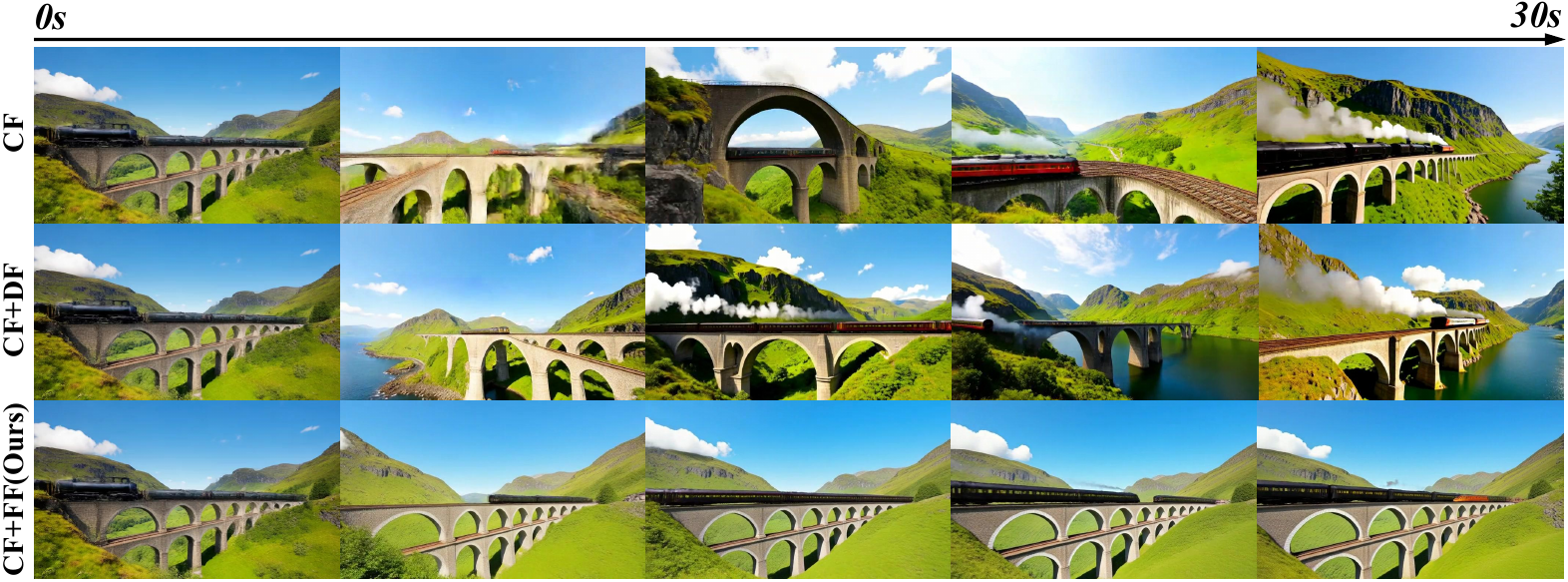}
    \caption{Additional visualization results for Causal-Forcing in 30-second video generation.}
    \label{fig:visual_appendix_30s_cf}
\end{figure}

\begin{figure}[ht]
    \centering
    \includegraphics[width=\linewidth]{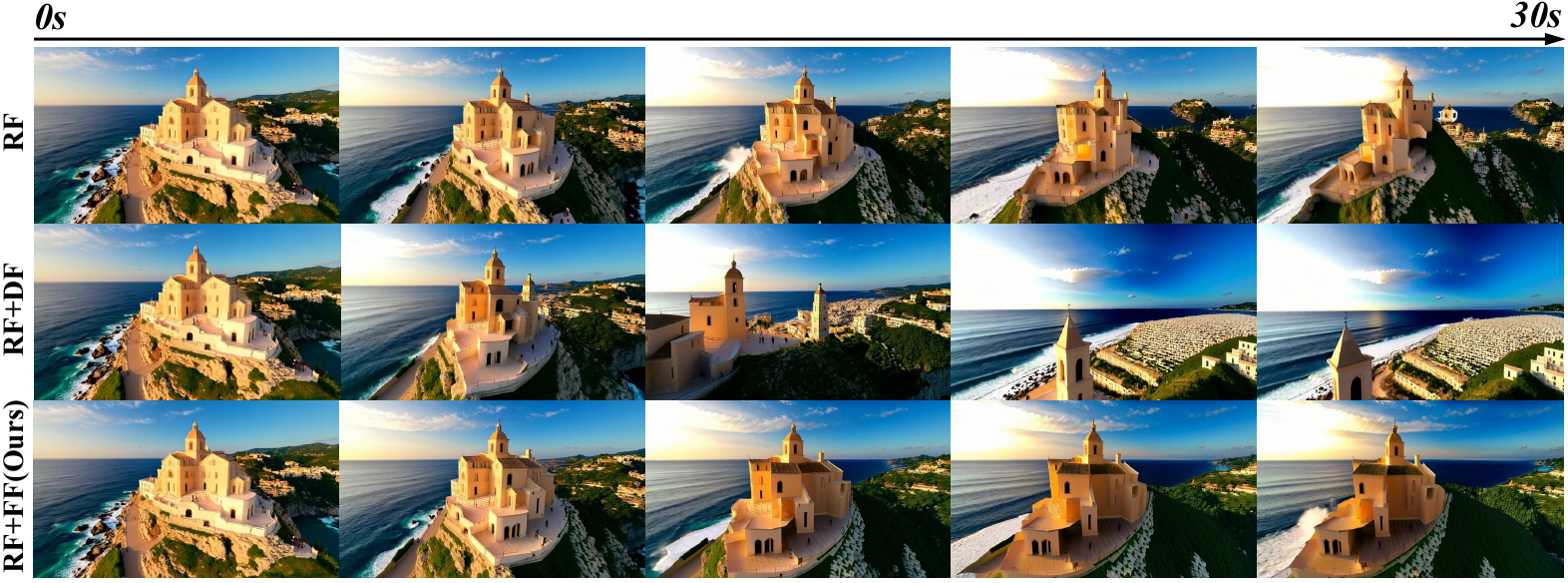}
    \caption{Additional visualization results for Reward-Forcing in 30-second video generation.}
    \label{fig:visual_appendix_30s_rf}
\end{figure}

\begin{figure}[ht]
    \centering
    \includegraphics[width=\linewidth]{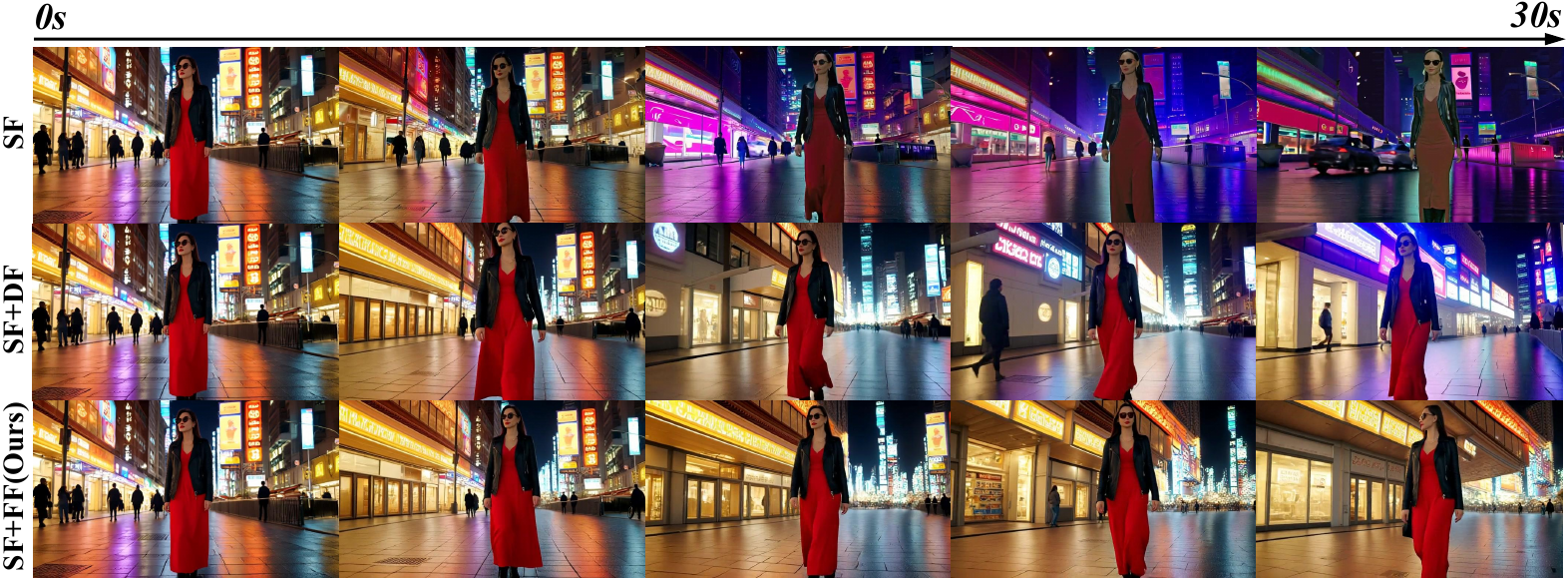}
    \caption{Additional visualization results for Self-Forcing in 30-second video generation.}
    \label{fig:visual_appendix_30s_sf}
\end{figure}

\begin{figure}[t]
    \centering
    \includegraphics[width=\linewidth]{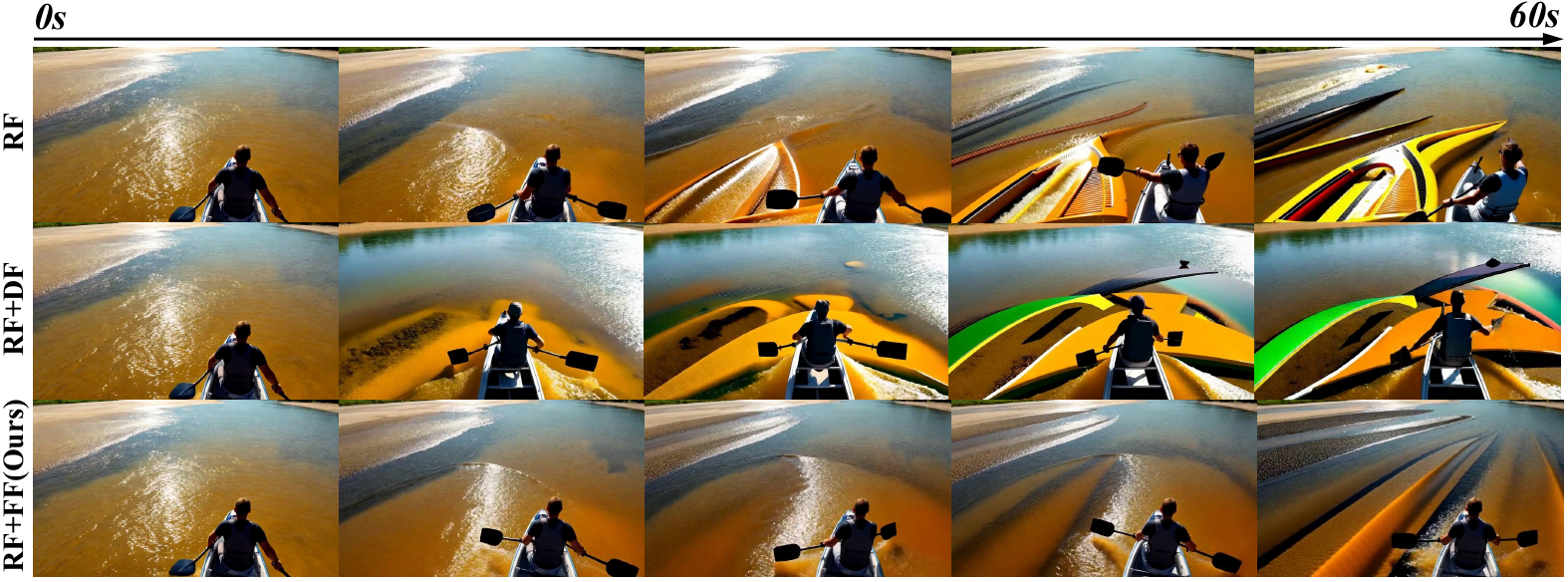}
    \vspace{-1em}
    \caption{Additional visualization results for Reward-Forcing in 60-second video generation.}
    \vspace{-1em}
    \label{fig:visual_appendix_60s_rf}
\end{figure}

\begin{figure}[t]
    \centering
    \includegraphics[width=\linewidth]{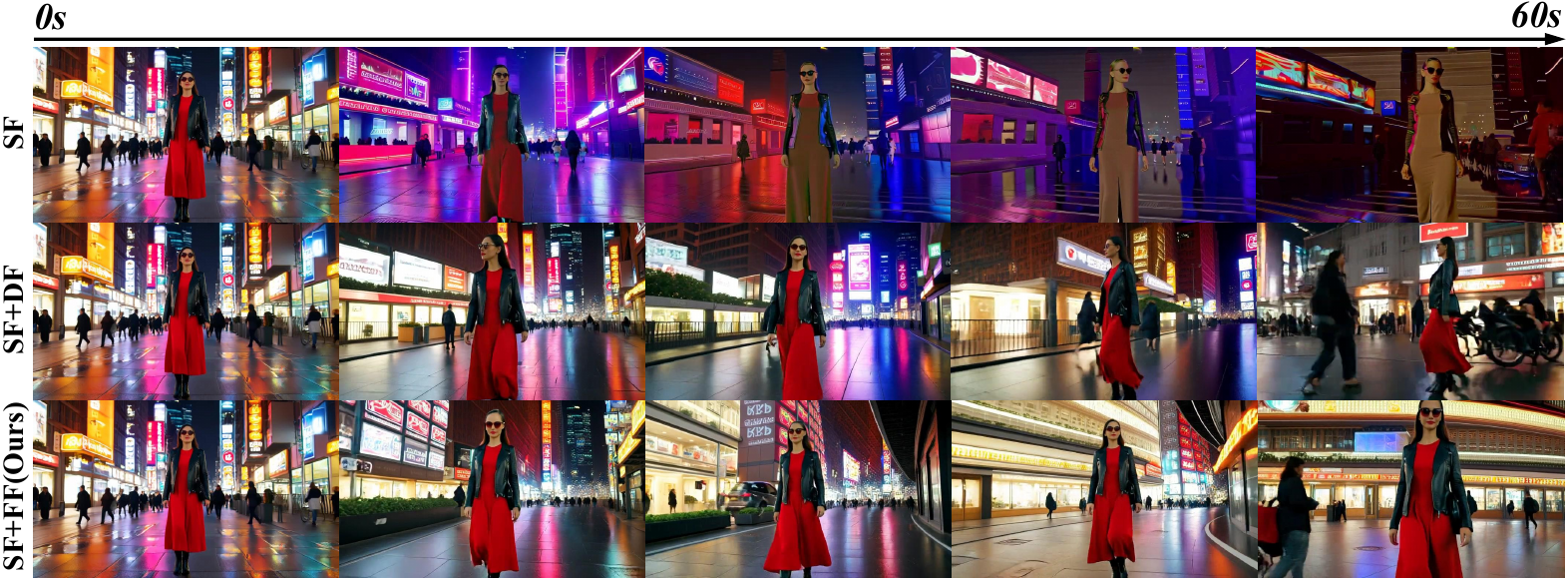}
    \vspace{-1em}
    \caption{Additional visualization results for Self-Forcing in 60-second video generation.}
    \vspace{-1em}
    \label{fig:visual_appendix_60s_sf}
\end{figure}

To further facilitate visual comparison, we provide additional qualitative results comparing \modelname\ with representative baselines in Figure~\ref{fig:visual_appendix_30s_cf}, Figure~\ref{fig:visual_appendix_30s_rf}, Figure~\ref{fig:visual_appendix_30s_sf}, Figure~\ref{fig:visual_appendix_60s_rf}, and Figure~\ref{fig:visual_appendix_60s_sf}.

\textbf{Discussion about the dynamic degree.} 
For the Causal-Forcing backbone, \modelname\ obtains a lower dynamic degree than some baselines, as shown in Table~\ref{tab:long-video} and Figure~\ref{fig:visual_appendix_30s_cf}. 
This does \textbf{not} indicate degraded motion dynamics; rather, the higher dynamic degree of the native KV-cache strategy and Causal-Forcing+Deep-Forcing often comes from \textbf{abrupt, undesired scene changes} during unstable long-horizon generation, rather than meaningful object or camera motion. In contrast, \modelname\ avoids such transitions and achieves \textbf{reasonable and stable dynamics with better long-horizon consistency}, as shown in Figure~\ref{fig:visual_appendix_30s_rf}, Figure~\ref{fig:visual_appendix_30s_sf}, Figure~\ref{fig:visual_appendix_60s_rf}, Figure~\ref{fig:visual_appendix_60s_sf}, and our anonymous website.

\section{Quantitative Analysis of Pre-RoPE Query Stability}
\label{sec:appendix_stability_quantitative}
To further quantify the cross-frame stability of pre-RoPE queries, we compute the average pairwise Wasserstein distance between frame-wise query distributions. To avoid scale-dependent comparisons, we normalize the distance by the pooled standard deviation and report normalized \(W_1\). Lower values indicate stronger cross-frame distributional stability. As shown in Table~\ref{tab:normalized_w1_query_stability}, pre-RoPE queries consistently exhibit much smaller normalized \(W_1\) than RoPE-modulated queries across different autoregressive video generation backbones, query dimensions, and prompts. Across all evaluated settings, pre-RoPE queries achieve an average normalized \(W_1\) of only \(0.123\sigma\), whereas RoPE-modulated queries reach \(1.176\sigma\), showing nearly one order of magnitude larger distributional drift. 
\begin{table}[H]
\centering
\caption{Normalized pairwise Wasserstein distance \(W_1\) of pre-RoPE and RoPE-modulated queries.}
\label{tab:normalized_w1_query_stability}
\resizebox{\linewidth}{!}{
\begin{tabular}{llcccccc}
\toprule
\multirow{2}{*}{Backbone} & \multirow{2}{*}{Dim.} 
& \multicolumn{3}{c}{Pre-RoPE Query} 
& \multicolumn{3}{c}{RoPE-Modulated Query} \\
\cmidrule(lr){3-5} \cmidrule(lr){6-8}
& & Prompt 0 & Prompt 1 & Prompt 2 & Prompt 0 & Prompt 1 & Prompt 2 \\
\midrule
\multirow{2}{*}{Self Forcing} 
& 0 & 0.098 & 0.122 & 0.204 & 1.192 & 1.192 & 1.192 \\
& 5 & 0.092 & 0.092 & 0.094 & 1.187 & 1.187 & 1.188 \\
\midrule
\multirow{2}{*}{Causal Forcing} 
& 0 & 0.139 & 0.147 & 0.152 & 1.190 & 1.188 & 1.191 \\
& 5 & 0.103 & 0.103 & 0.150 & 1.185 & 1.185 & 1.185 \\
\midrule
\multirow{2}{*}{LongLive} 
& 0 & 0.126 & 0.113 & 0.207 & 1.192 & 1.193 & 1.194 \\
& 5 & 0.097 & 0.167 & 0.059 & 1.187 & 1.188 & 1.189 \\
\midrule
\multirow{2}{*}{Reward Forcing} 
& 0 & 0.126 & 0.088 & 0.195 & 1.149 & 1.148 & 1.150 \\
& 5 & 0.066 & 0.101 & 0.068 & 1.110 & 1.086 & 1.102 \\
\midrule
\multirow{2}{*}{Rolling Forcing} 
& 0 & 0.168 & 0.083 & 0.206 & 1.192 & 1.194 & 1.192 \\
& 5 & 0.093 & 0.093 & 0.130 & 1.187 & 1.189 & 1.186 \\
\bottomrule
\end{tabular}
}
\end{table}

\section{Pre-RoPE Stability under Highly Dynamic Scenarios}
\label{sec:highly_dynamic_appendix}
To further examine whether pre-RoPE query stability holds under highly dynamic visual conditions, we construct three challenging prompts involving rapid camera motion, complex object interactions, and substantial scene dynamics:

\begin{tcolorbox}[
    colback=gray!5,
    colframe=gray!45,
    boxrule=0.6pt,
    arc=2pt,
    left=6pt,
    right=6pt,
    top=6pt,
    bottom=6pt,
    title=\textbf{Highly Dynamic Prompts}
]
\begin{enumerate}[leftmargin=*]
    \item A drone-level chase through a neon night market, weaving between sprinting cyclists, dancing crowds, flying paper lanterns, and sudden rain as the camera whips around tight corners.
    \item A storm rescue on a rocky coast, with a helicopter spotlight sweeping across crashing waves, a lifeboat pitching violently, lightning flashes, spray hitting the lens, and people moving in several directions.
    \item A rooftop parkour race at sunset, jumping over vents and glass skylights while trains pass below, birds scatter, shadows stretch fast, and the camera swings from wide aerial views to close handheld pursuit.
\end{enumerate}
\end{tcolorbox}

\begin{figure}[H]
    \centering
    \begin{subfigure}[t]{0.45\linewidth}
        \centering
        \includegraphics[width=\linewidth]{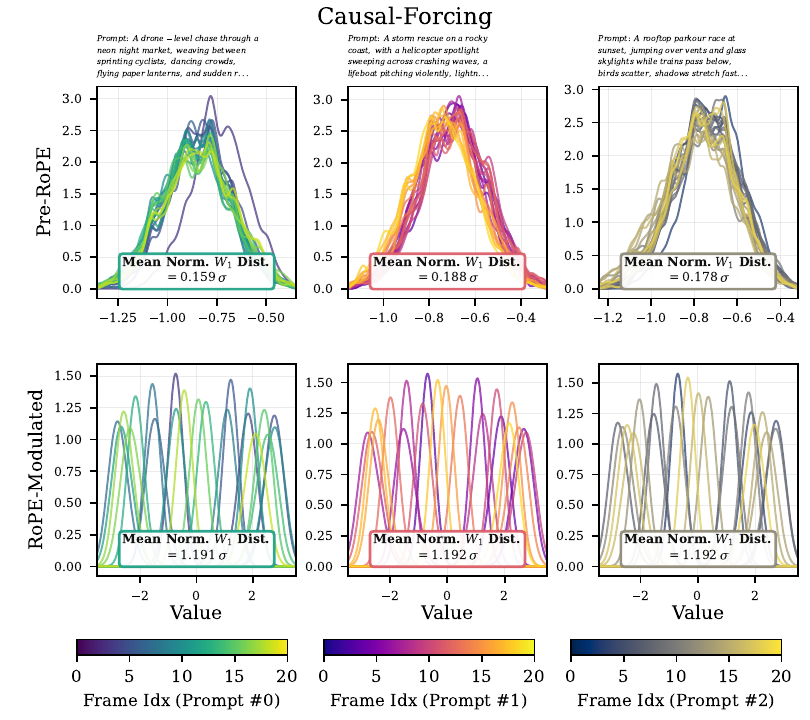}
        \caption{Causal Forcing}
        \label{fig:dynamic_high_motion_causal}
    \end{subfigure}
    \hfill
    \begin{subfigure}[t]{0.45\linewidth}
        \centering
        \includegraphics[width=\linewidth]{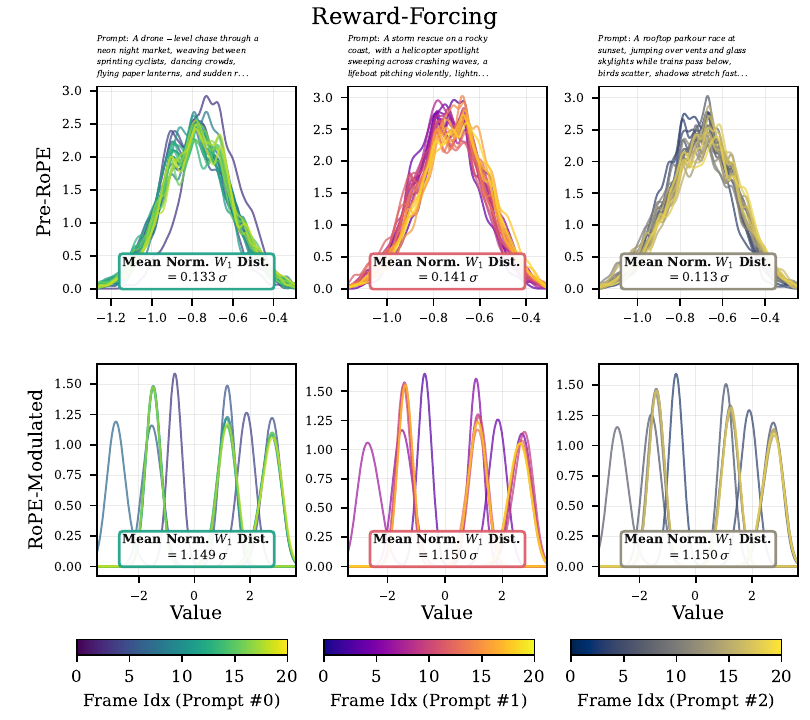}
        \caption{Reward Forcing}
        \label{fig:dynamic_high_motion_reward}
    \end{subfigure}

    \begin{subfigure}[t]{0.45\linewidth}
        \centering
        \includegraphics[width=\linewidth]{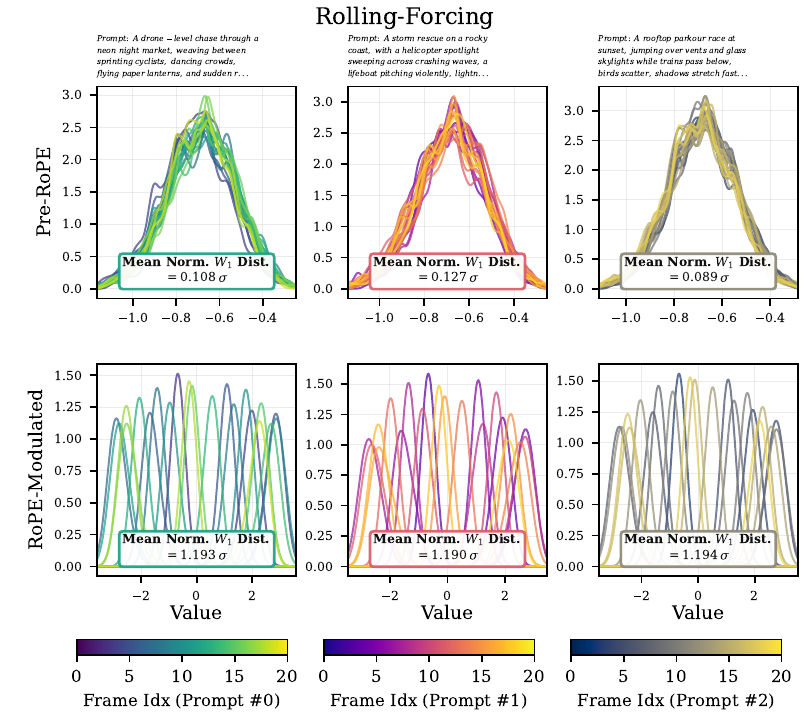}
        \caption{Rolling Forcing}
        \label{fig:dynamic_high_motion_rolling}
    \end{subfigure}
    \hfill
    \begin{subfigure}[t]{0.45\linewidth}
        \centering
        \includegraphics[width=\linewidth]{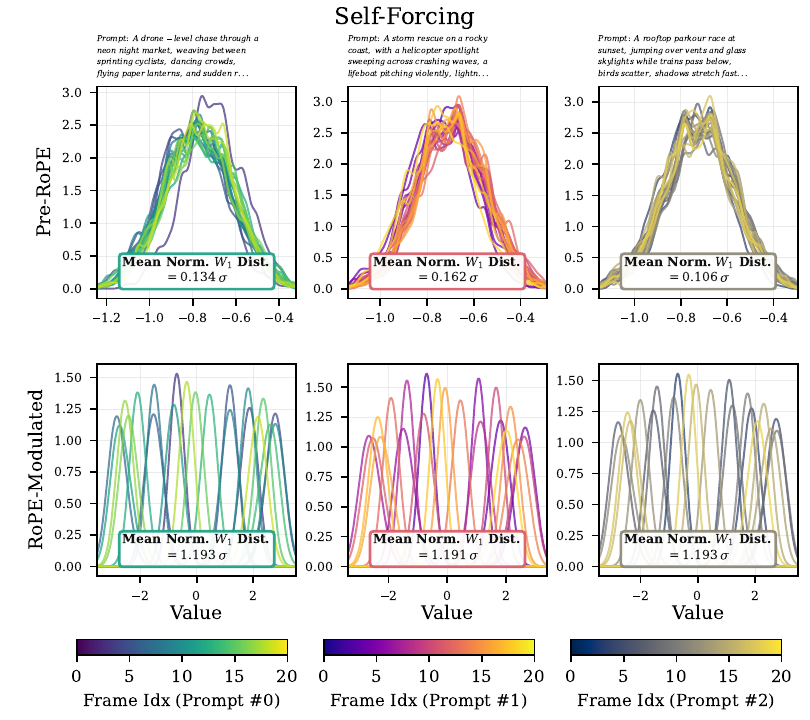}
        \caption{Self Forcing}
        \label{fig:dynamic_high_motion_self}
    \end{subfigure}

    \caption{Pre-RoPE and RoPE-modulated query distributions under \textbf{highly dynamic scenarios} across representative autoregressive video generation models.}
    \label{fig:dynamic_high_motion_query_distribution}
\end{figure}

We visualize the distributions of pre-RoPE queries and RoPE-modulated queries under these scenarios in Figure~\ref{fig:dynamic_high_motion_query_distribution}. The results show that even under strong motion, fast viewpoint changes, and dense dynamic interactions, the canonical pre-RoPE query distributions remain substantially more stable than their RoPE-modulated counterparts.

\clearpage
\newpage

\end{document}